\documentclass{article}
\usepackage[pdftex]{graphicx}
\usepackage{gensymb}
\usepackage{url}
\DeclareGraphicsExtensions{.pdf,.jpeg,.png,.jpg}
\usepackage{amsmath}
\usepackage{array}
\usepackage{verbatim}     
\usepackage{siunitx} 
\usepackage{multirow}
\usepackage{booktabs}
\begin{document}

\title{Solving Visual Object Ambiguities when Pointing: An Unsupervised Learning Approach
}


\author{Doreen Jirak*\thanks{University of Hamburg, Germany}         \and
        David Biertimpel*\thanks{University of Amsterdam, Netherlands} \and 
Matthias Kerzel\footnotemark[1]  \and
Stefan Wermter\footnotemark[1] 
}



\maketitle

\begin{abstract}
Whenever we are addressing a specific object or refer to a certain spatial location, we are using referential or deictic gestures usually accompanied by some verbal description. Especially pointing gestures are necessary to dissolve ambiguities in a scene and they are of crucial importance when verbal communication may fail due to environmental conditions or when two persons simply do not speak the same language. With the currently increasing advances of humanoid robots and their future integration in domestic domains, the development of gesture interfaces complementing human-robot interaction scenarios is of substantial interest. The implementation of an intuitive gesture scenario is still challenging because both the pointing intention and the corresponding object have to be correctly recognized in real-time. The demand increases when considering pointing gestures in a cluttered environment, as is the case in households. Also, humans perform pointing in many different ways and those variations have to be captured. 
Research in this field often proposes a set of geometrical computations which do not scale well with the number of gestures and objects, use specific markers or a predefined set of pointing directions.
In this paper, we propose an unsupervised learning approach to model the distribution of pointing gestures using a growing-when-required (GWR) network. We introduce an interaction scenario with a humanoid robot and define so-called ambiguity classes. Our implementation for the hand and object detection is independent of any markers or skeleton models, thus it can be easily reproduced. Our evaluation comparing a baseline computer vision approach with our GWR model shows that the pointing-object association is well learned even in cases of ambiguities resulting from close object proximity.


\end{abstract}

\section{Introduction}
\label{intro}
Deictic gestures are the nonverbal complementary hand actions to the linguistic \textit{deixis}, which are used when referring to an object, a person, or a spatial location. Similar to a word or sentence, deictic gestures cannot be fully understood without the context. Especially when verbal communication is hampered by environmental conditions or fails completely, pointing gestures can help to dissolve ambiguities by \textit{pointing} to the meant person, object or area. A special area of interest to study the impact of pointing gestures on learning higher cognitive tasks like object recognition and language acquisition is developmental psychology \cite{Carpe_98}\cite{Tomas_07}\cite{Behne_12}\cite{Kishi_17}. Intuitively, pointing to an object is usually accompanied by labeling it correspondingly,
typically in an infant-parent setting, e.g.: ``\textit{This} is your \textit{teddy bear}" (parent points to an object) or ``Are you \textit{looking for} your \textit{teddy bear}?" (infant points to an object or object location).
Eventually, deictic gestures have been shown to facilitate joint attention and synchronization processes for robot-robot interaction \cite{Kyrki_15}\cite{Kyrki_19}\cite{Gromo_18}. 

The development of deictic gesture interfaces comprises the following issues on the study design: the concrete definition of deictics, choice of sensors, conductance of a human-robot interaction (HRI) study delivering possible explanations for understanding gestures (including verbal communication) or proposing learning strategies for robotic operations. We provide a literature review based on the described issues, followed by our description of the experimental settings and the introduction of the humanoid robot platform employed in our study. We give descriptions of the computer vision approach used to extract significant features of pointing gestures including our model for a users' pointing intention. We also explain how to model pointing gestures using growing-when-required networks (GWR), an unsupervised learning approach capturing the distribution of deictics. We present the results of a comparative evaluation of both methods and close the paper with a discussion and indications of potential future application of our proposed framework. 

\subsection{Related Work}
The development of natural interaction systems is important to foster communicative strategies between humans and robots and to increase their social acceptance \cite{Salem_13}. Deictic gestures allow the establishment of joint attention in a shared human-robot workspace for cooperative tasks and they complement or even substitute verbal descriptions in spatial guidance or when referring to specific objects. 
The following studies were conducted on distinct robot platforms and employ different vision capture systems.

In \cite{Hagit_07}, the authors evaluated an HRI scenario between the humanoid robot `Robovie' and a human subject was presented . A participant was asked to guide the robot through a package stacking task. The authors put emphasis on attention synchronization when initializing the interaction, the context and the `believability' of the whole scenario, subsumed under the term \textit{facilitation process}. The underlying hypothesis was that the implementation of this process supports the naturalness of the interaction between the human and the robot. Hence, the whole experiment was driven by two conditions, namely the instruction method using deictics or symbolic expressions (i.e. asking for a box identification number) and whether or not there would be some facilitation factor during the guidance process, as the robot was able to ask for some error correction. Subjects were equipped with a motion capturing system to record the pointing gestures, and an additional speech interface was provided for verbal instruction. In total, 30 persons participated in the study and their responses to the overall scenario regarding six options like `quickness' and `naturalness' were recorded. Overall, the ANOVA test statistics revealed positive enhancement of the interaction when the HRI scenario included the facilitation process, which might be explained by the fact that gestures support verbal communication. Also, the additional feedback made by the robot may strengthen the cooperative nature of the task.

Although deictic gestures are primarily used to represent pointing gestures, Saupp\'{e} and Mutlu \cite{Saupp_14} defined a set of six deictic gestures types, each being used to refer either to a single object or object clusters (specific area of multiple objects). Specifically, the authors distinguished between `pointing', `presenting', `touching', `exhibiting', `grouping', and `sweeping' gestures. The whole experiment setup comprised speech with two conditions (full and minimal description), gaze and the named gestures. The humanoid NAO robot was selected as the interaction partner, trained to produce each gesture by a human teacher beforehand (i.e. the motor behavior was executed by a human on the robot). The on-board RGB camera was used as the vision sensor capturing the scene. The objects were differently colored wooden toys with rectangle, square and triangular shape.
Experiments with 24 subjects  were conducted employing different conditions like noise and ambiguity as well as the neutral condition providing the simplest and clearest setup. An ANOVA evaluation from questionnaires  was driven by the question of communicative expressiveness when including gestures and their effectiveness. In essence, the study revealed significant effects of gestures, where `exhibiting' and `touching' stood out in both expressiveness and effectiveness. Both gesture types involve acting on the object, which directly solves ambiguities and visibility issues, especially when lifting a desired object. This result is conclusive when considering a scenario, e.g. in a shop, when both the customer and the vendor do not speak the same language and hence any verbal descriptions become obsolete. However, in a concrete robotic implementation this extra step can be critical in real-time applications or when the computational capacity is low, as also pointed out by the authors of the study. In addition, larger distances between a robot arm and the requested object hinder a meaningful execution of deictic gestures. The `pointing' gestures performed worse than the other gestures except when the distance between the subject and the robot was large and humans presumably use pointing in those cases to refer to a specific object area. 

To investigate object ambiguities resulting from different pointing directions, Cosgun et al. \cite{Cosgu_15} implemented both a simulation and a real experiment using an industrial robot arm. The Kinect camera was employed as the necessary vision modality, delivering object point clouds which were used to determine the degree of intersection whenever a pointing gesture with one arm was performed. The latter was represented as a set of joint angles between head-hand and head-elbow computed from the OpenNI skeleton tracker. Six subjects performed pointing in seven target directions, which yielded the horizontal and vertical angles. In a subsequent error analysis between the intended and the ground truth pointing target, the Mahalanobis distance to the objects was computed.
For the real experiments, two cans were positioned central in the sensor plane and a subject new to the experiment performed pointing accordingly. The distance between the two objects was varied where 2cm was the minimum. The authors reported a good separation score for object distances larger than $3cm$ while for lower values the system detected ambiguity. Since the approach uses simple methods to represent pointing gestures, the approach is limited in the flexibility of pointing positions as the targets on the table were defined beforehand. Also, the concrete evaluation falls short because only two objects were arranged in one row without any ambiguities.

A similar scenario using also a robot arm and the Kinect but utilizing multiple objects was presented in \cite{Shukl_15}. First, synthetic images of hand poses from varying viewpoints were used to extract important features used to train a probabilistic model. This model was used to obtain training and test distributions, where only for the training set the viewpoint is known. Second, a similarity score based on cross-correlation between the two distribution was calculated between train and test images after correcting the latter for any transformations due to the distinct viewpoints. All poses were represented in 6D including the direction from which a distance to objects was estimated. All possible distances were calculated for all objects and evaluated using a confidence measure. The object associated with the highest value was then identified as the pointing target. In total, ten objects from the YCB dataset were used, known to be appropriate objects for robot manipulation tasks. The authors \cite{Shukl_15} created a dataset comprising 39 training images from object configurations with distinct horizontal and vertical angles. 180 images showing different gesture scenarios were used for evaluation, 12 for tuning the incorporated kernel in the model and the remaining 168 for testing with nine subjects. The pointing was further split into hand poses from the subject and using an additional tool to enhance the pointing direction. Both the mean and the standard deviation for the best pointing direction estimate were computed as well as the best nearest estimate. The approach showed good qualitative performance for both pointing conditions, which means that there might be no differences between using only the hand or performing the pointing with a tool. Also, the model demonstrated robustness for the differently shaped hands by the participants. However, when the objects were less than $10\degree$ apart (the proximity values used for training) the analysis revealed object ambiguities, leading eventually to incorrect associations between pointing direction and desired object.
In a quantitative evaluation, the model achieved best performance of 99.40\% of accuracy when the acceptable error range was $[-15\degree; 15\degree]$ and an additionally introduced factor $t=1.0$, $t \in[0;1]$, which weights the hypothesis scores returned by the estimation procedure. For lower error ranges, the accuracy dropped significantly, e.g. to about 55\% for 5\degree and 20\% for about 3\degree. 

A study related to explain developmental stages in infants understanding deictics for joint attention was presented by Y. Nagai \cite{Nagai_05}. An HRI scenario using a humanoid robot called `Infanoid' was established integrating pointing gestures with an object which was pointed to: pointing with the index finger or using the whole hand (referred to as reaching) and tapping. The implementation was based on capturing images and extracting significant edge features of the pointing hand, as well as optical flow \cite{Nagai_05a}. Those features were learned with a neural network architecture, predicting the corresponding motor commands to guide the robots' gaze direction. The evaluation in a one-person setup showed superior learning performance for reaching compared to pointing for the edge images. Using optical flow, learning the correct gaze direction was superior for tapping, followed by pointing with movement and, last, pointing without movement. These results underpin the hypothesis that deictic comprehension in infants is facilitated by moving gestures \cite{Moore_97}\cite{Astor_19}. 

Canal et al. \cite{Canal_16} integrated pointing gestures in different settings outside the lab (high school, community center and elderly association) and evaluated their system including participants of broad age ranges (9-86) is demonstrated. Again, the Kinect device was used to compute significant body joints representing the pointing and waving gesture from the delivered skeleton model. The dynamic nature of the gesture was captured by dynamic time warping (DTW), aligning two time series to determine their degree of similarity. When a pointing was detected, the PCL library was used for ground plane extraction and the corresponding direction was estimated using the hand-elbow coordinates. The authors emphasized a limitation for the concrete pointing due to sensor limitation of the Kinect itself, which made it necessary to point with the arm in front of the body. Also, only an object detection routine was implemented to check the presence of an object. Three objects were used, two milk bottles and a cookie box. Possible object ambiguities were resolved by a robots' reply to the pointing gesture, requesting the users' help for disambiguation, e.g. asking whether the left or right object was desired. 67 subjects unfamiliar with robots tested the HRI scenario and documented their experience in a survey. The questionnaire asked for the response time of the gestures, the accuracy of the pointing and the naturalness of the whole interaction, evaluated on a scale 1-5 (worst-best). While the participants rated the response time relatively high across the three age groups, the concrete object detection with the pointing was more nuanced, showing a decrease in approval in the groups 35-60 and 61-86. This result is in agreement with the evaluation of the object detection alone, which yielded a recognition rate of 63.33\%. This is comparatively low when taking into account a recognition rate of 96.67\% for the pointing gesture. Interestingly, the naturalness of the interaction was rated highest in the group of the elderly, which are a target audience when it comes to service robots and assisting systems in health care.  

\subsection{Research Contribution}
The results obtained by the studies described above are important for the development of robotic systems in domestic and health care systems, set out for human-robot interaction (HRI). However, most approaches rely on skeleton models obtained by depth sensors, which are not implemented in common robot platforms and thus need an additional hardware setup. This, in turn, can introduce more noise and stability problems when the hardware is e.g. equipped on the robot head. Also, until today only a few studies address object ambiguities in interaction scenarios but concentrate on the different performance of pointing gesture or deictics in general.

In the light of successfully benchmarking
gesture and object recognition, predominantly using supervised deep learning implementations, using an unsupervised learning strategy does not seem to be intuitive in the first place.
However, learning gesture-object associations includes different modeling stages where deep learning, due to its dependence on large training data, has no or only minimal benefit and, thus, does not always qualify to be the potentially best approach in robotics \cite{Jirak_17}. Hence, we propose our approach using both traditional computer vision and, complementary, introduce the Growing-When-Required network (GWR, \cite{Marsl_02}) as an unsupervised approach supporting pointing gesture recognition even in ambiguous object arrangements.

\section{Experiment Layout}
In this section, we outline the interaction pointing scenario between the NICO robot and a human participant, explain our design principles as well as
the data recordings. We also introduce our definition of \textit{ambiguity classes} with different level of difficulty. 

\subsection{The NICO robot}

The NICO robot is a humanoid robot which serves as a fully controllable robot for multimodal human-robot interaction scenarios. NICO is an abbreviation for ``Neurally-Inspired Companion", signifying its integration into assisting tasks like object grasping \cite{Hafez_19} and serving as a reliable platform for HRI studies including emotion recognition \cite{Chura_17} and human-robot dialogue understanding \cite{Sique_18} as well as for the assessment of critical design factors for robots e.g. ``likeability" or safety issues \cite{Kerze_17}, which are crucially important aspects for social robot research. The robot architecture and API is configured in a modular fashion to facilitate the integration, extension, and combination of task-specific algorithms. Our implementation follows this specification and is either ready for use as a standalone software package also for other robot platforms or complements more complex HRI scenarios.
A full description of the robot hardware and details on the software are available in \cite{Kerze_17}.

\subsection{Experiment Design}
\begin{figure}[thpb]
       \centering
        \includegraphics[width=0.9\textwidth]{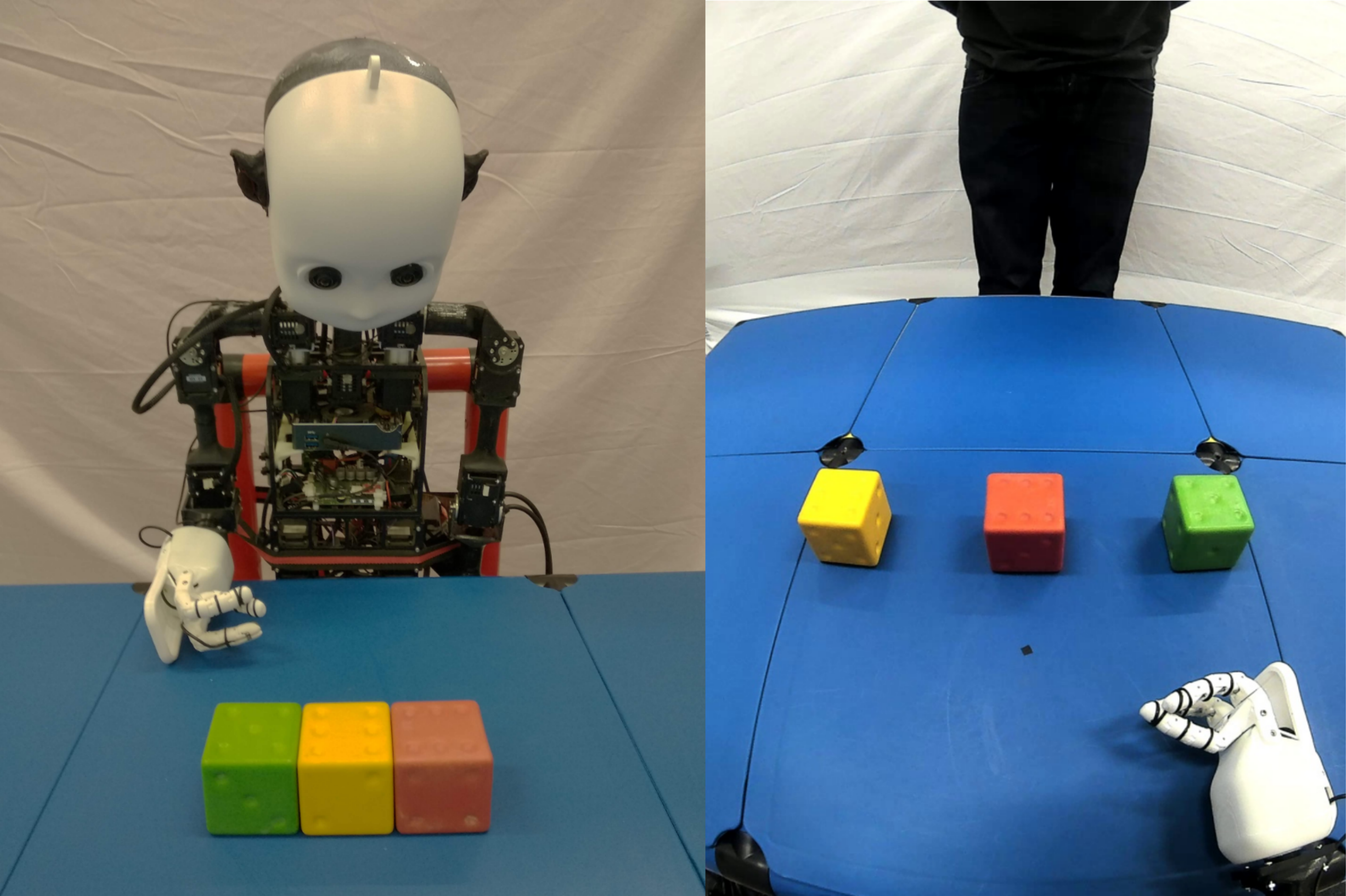}
       \caption{Scenario outline from the perspective of the participant (left) and from the perspective of the NICO robot (right).}
        \label{fig:nico}
\end{figure}

For our pointing scenario, we defined a controlled experimental setup to reduce noise introduced by environmental variability. In particular, we constrained the interaction to happen only between the NICO robot and one participant in the scene facing each other. NICO is seated opposite to the human participant in a typical face-to-face configuration that could, for instance, occur in an instruction setting.
We fixed both the illumination conditions in the room and the background, the latter restricted to white curtains. The pointing videos were always recorded from the same perspective using the camera available in the NICO head. Besides the visual recording of the deictic gestures, NICO serves as an interaction partner. NICO's cameras are embedded into a human-like head, and its direction of gaze and field of view can be assumed by the human interaction partner, allowing him or her to adjust the deictic gestures according to the HRI situation.
As we focus on the pointing gesture, we chose three differently colored cubes on a blue desk to facilitate the object detection and recognition process in our system. We employ a maximum of three objects to keep the number of object permutations and the ambiguity classes manageable.   
The participant performs the pointing with the right hand and we only recorded the upper body and arms as we did not take any facial expressions into consideration. Due to a fixed standing position, the pointing hand enters NICO's vision center during the gesture performance. The viewpoints in the scenario are demonstrated in Figure \ref{fig:nico}. While NICO is not depicted in the recorded images, it is the intended recipient of the deictic gestures performed by the human participant.

\begin{figure}[thpb]
       \centering
        \includegraphics[width=0.9\textwidth]{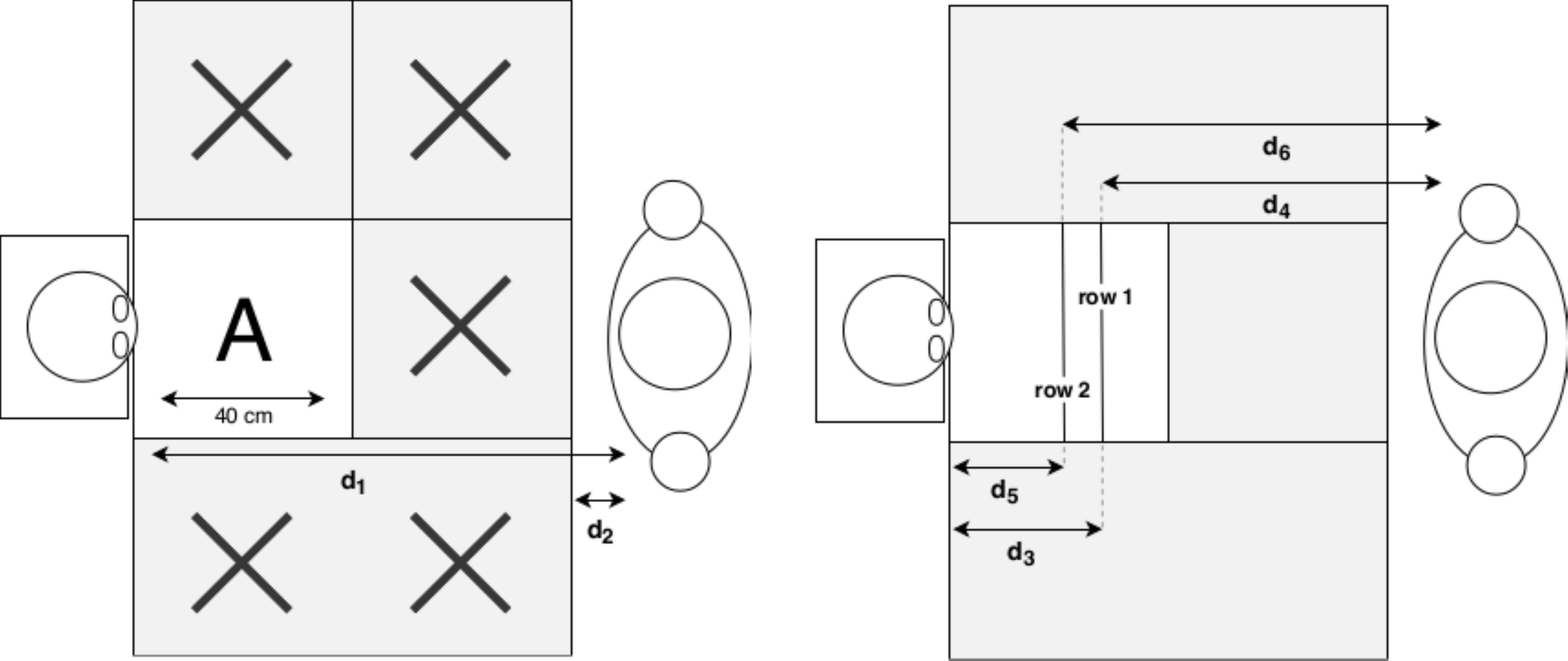}
       \caption{}{Outline of the object area A, and the defined distances and rows for the pointing scenario.}
       \label{fig:distances}
\end{figure}

Having explained the general setting, we will now proceed with an outline of the specific table arrangements.
The fixed positions of both, the NICO robot and the human subject, allowed us to measure some distances which become important in the subsequent computations of the hand and pointing detection as well as the object arrangements. A complete picture of all distance measurements is provided in Figure \ref{fig:distances}.

First, we measured the distance between the NICO robot and the human participant, which we call $d_1=100$cm. The distance between the robot and the table is $0$cm (the robot is directly sitting at the table) while the experimenter leaves a gap of $d_2=20$, which is a trade-off between the human anatomy and the natural distance in an interaction. 
The table area $A=40 \times 40$ cm is the space for possible object arrangements. The three objects used in our experiments are uniformly sized cubes with an edge length of $5.8$cm. We defined \textit{row1} and \textit{row2}, from where the objects can be placed in various positions. This setup also limits the search space for pointing gestures, which is necessary to evaluate our system.

The \textit{row 1} is $28$cm away from the NICO robot, denoted as $d_3$, and $65$cm away from the experimenter, labeled as $d_4$. To define \textit{row 2}, we measured the distance $d_5=21$cm from the robot and $d_6=72$cm from the human subject.
From the robot perspective, \textit{row1} is located above \textit{row2}.

\subsection{Data Collection and Preprocessing}

\begin{figure}[thpb]
       \centering
        \includegraphics[width=0.95\textwidth]{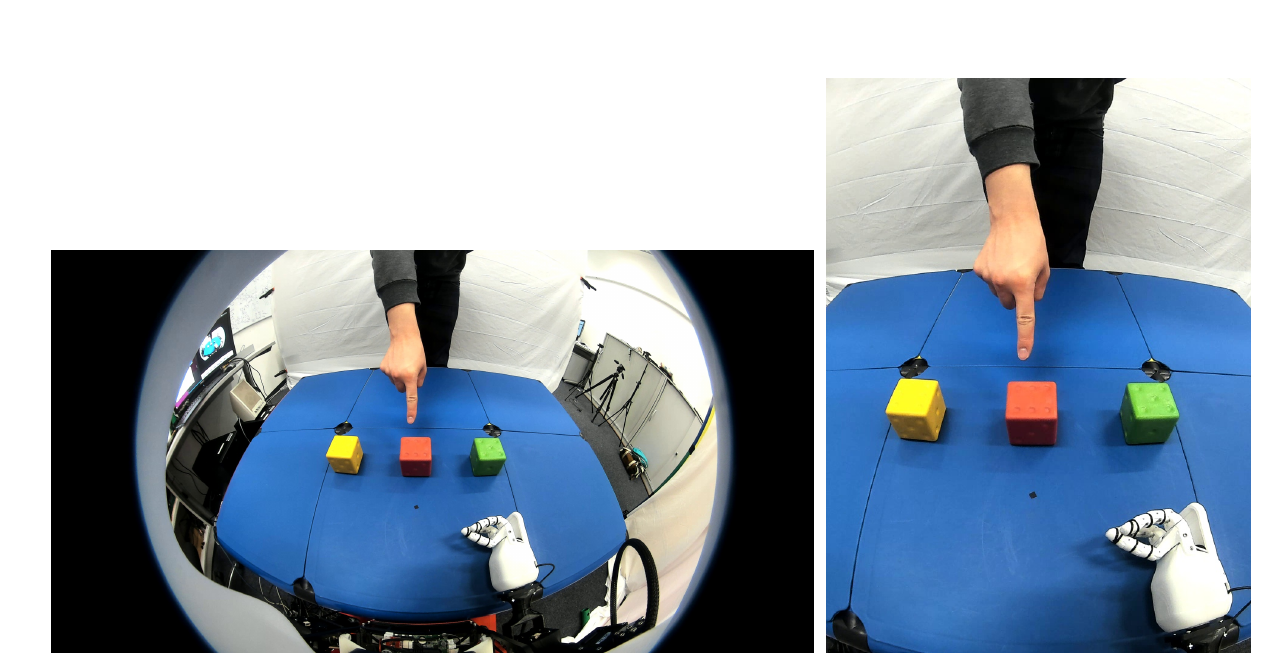}
       \caption{Left: Image caption with the fish-eye lens installed in the NICO robot. Right: The image after applying the preprocessing step.}
       \label{fig:fisheye}
\end{figure}

We collected image data from the right NICO RGB camera (initial resolution: $4096\times 2160$ pixels). To erase effects introduced by the fisheye lens, we scaled the recorded images down to $2048\times 1080$ and cropped them to $700\times 900$ to exclude unnecessary background information. 
Figure \ref{fig:fisheye} demonstrates the effect of the procedure.

\subsection{Ambiguity Classes} 
\begin{figure}[]
       \centering
        \includegraphics[width=0.55\textwidth]{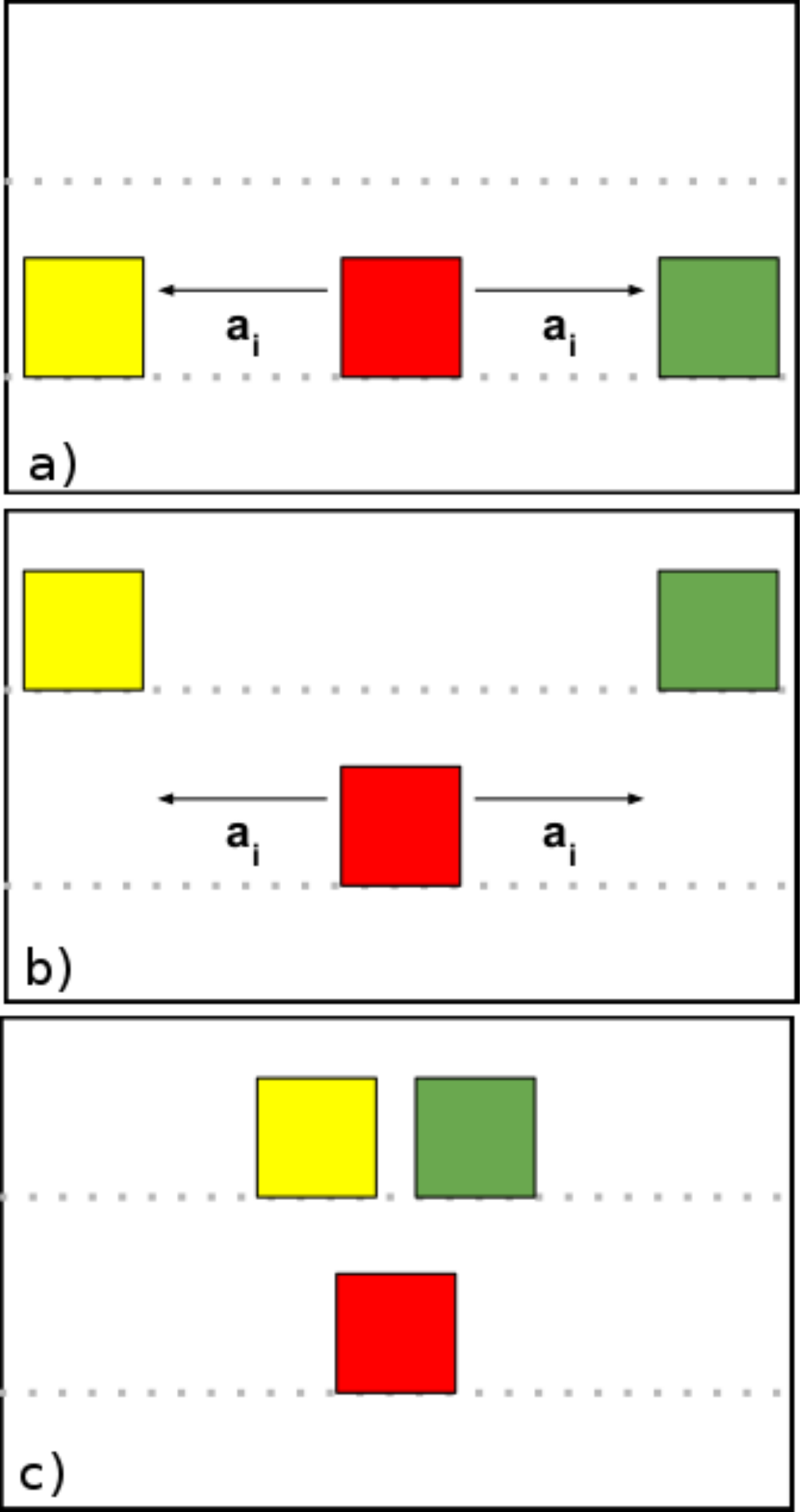}
       \caption{Layout of the different ambiguity classes. a) Objects are aligned on one row, the distances between the objects are specific for each ambiguity class $a_i$. b) Three objects distributed on the two rows defined for the scenario. c) Layout for ambiguity class $a_4$. Due to the perspective of the robot, the object on the lower row has an overlap with the other two objects in the top row.}
       \label{fig:ambig_classes}
\end{figure}

To evaluate the correctness of pointing gestures with an associated object for different levels of ambiguity, we introduce four ambiguity classes $a_1-a_4$. Specifically, we consider the \textit{row} definitions shown in Figure \ref{fig:distances} and group the objects either into one row (\textit{row1} or \textit{row2}) or arrange them into a V-shape, as demonstrated in Figure \ref{fig:ambig_classes}. For increasing class index $a_i$ we reduced the distances between the objects:

\begin{itemize}
    \item ambiguity class: $a_1$: 10 cm
    \item ambiguity class: $a_2$:  5 cm
    \item ambiguity class: $a_3$:  0 cm
    \item ambiguity class: $a_4$:  overlapping
\end{itemize}

\begin{figure}[thpb]
       \centering
        \includegraphics[width=0.95\textwidth]{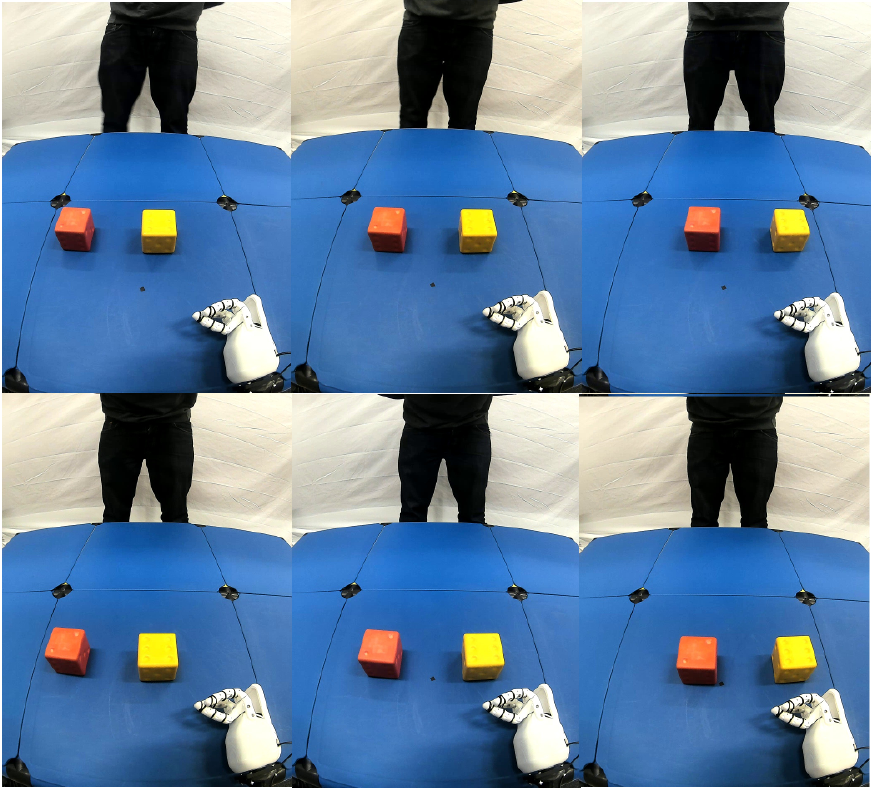}
       \caption{Exemplary demonstration of all possible object configurations on the defined rows for two objects in ambiguity class $a_1$.}
       \label{fig:a1}
\end{figure}

In total, we recorded 95 object configurations which are summarized in Table \ref{tab:ambig}. Note, that one object in a scene is unambiguous and thus not taken into consideration. The different numbers per ambiguity class $a_i$ result from the constraints of the distances. For instance, for ambiguity class $a_1$ the distance between two objects is 5cm, which results in six possible object arrangements on the two rows (shown in Figure \ref{fig:a1}). Correspondingly, three objects grouped on the table yields four possible configurations.

For the ambiguity class $a_4$, the objects are always distributed between \textit{row1} and \textit{row2}. From the robot's perspective, this arrangement results in an object overlap of $2 cm$, thus $a_4$ can be considered the class with the highest ambiguity.

 Additional 23 object combinations were recorded to include so-called ``edge cases" for a fair evaluation of the overall quality of our interface. The deictic gestures were recorded correspondingly. 


\begin{table}
\caption{Ambiguity Classes}
\label{tab:ambig}       
\begin{tabular}{lllllll}
\hline\noalign{\smallskip}
 & no ambig & $a_1$ & $a_2$ & $a_3$ & $a_4$ & $\sum$  \\
\noalign{\smallskip}\hline\noalign{\smallskip}
1 object & 16 & x & x & x & x & 16 \\ \hline
2 objects & x & 6 & 10 & 10 & 7 & 33 \\ \hline
3 objects & x & 4 & 12 & 10 & 20 & 46 \\ \hline
$\sum$ & 16 & 10 & 22 & 20 & 27 & 95 \\
\noalign{\smallskip}\hline
\end{tabular}
\end{table}

\section{Methodology}
Our work is driven by the goal to provide a natural and intuitive, yet fast deictic gesture interface. Our code is available for reproduction \footnote{\url{https://github.com/d4vidbiertmpl}}. We exclusively focus on vision-based techniques, i.e. we refrain from using any specifically colored gloves or markers and we also do not consider depth maps delivered by devices like the Kinect. The latter aspect is motivated by the fact that popular robot platforms either use stereoscopic vision or have only monocular cameras (see also \cite{Wu_12}), and we think that mounting another device is prone to introduce another error source. Also, only the upper body is involved in the gesture scenario, thus a whole-body representation including the calibration issues involved is unnecessary and not natural at all.

\subsection{Hand Detection}
We designed our system for HRI scenarios employing deictic gestures, thus the most important requirement for the interface implementation is to provide real-time processing of the hand detection and, subsequently, the hand segmentation. To achieve this, we follow some commonly known approaches in computer vision: first, we consider skin-color-based detection; hence we convert images from the RGB to YCbCr color space, keeping only the color distribution in the Cb and Cr components. Second, we choose 15 sample images from our recordings reflecting different hand shapes during pointing to objects, which provide different skin color distributions. To allow more flexibility in the hand detection, we decided to set up a skin color classifier which outputs either skin or no-skin color regions. 

For this purpose, we calculated the maximum likelihood estimates for a Cb, Cr pair to belong to skin color ($P(CbCr|skin)$) and not to belong to skin color ($P(CbCr|\sim skin)$). We created 2D histograms for both classes using the Cb and Cr components using pixel bins of ranges $[0;255]$. As a result, each histogram cell represents the pixel count of both components. We then convert each histogram into the respective conditional probability:

\begin{equation}
 P(CbCr|skin)= \frac{s[CbCr]}{T_s}   
\end{equation}

\begin{equation}
P(CbCr|\sim skin)= \frac{n[CbCr]}{T_n}
\end{equation}
where $s[CbCr]$ denotes the count of (CbCr) pairs in the skin histogram divided by the total count of histograms entries $T_s$, analogous to the non-skin part.

\begin{figure}[thpb]
       \centering
        \includegraphics[width=0.95\textwidth]{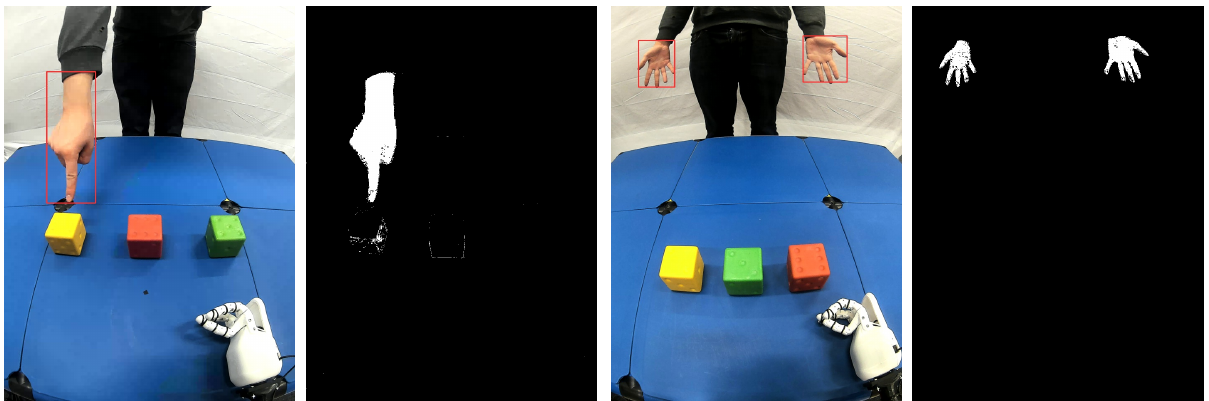}
       \caption{Demonstration of the resultant hand detection and hand segmentation. The images show that the segmentation works for different distances from the image plane (robot perspective).}
       \label{fig:handseg}
\end{figure}

We can then label a pixel as skin when:

\begin{equation}
 \frac{P(CbCr|skin)}{P(CbCr|\sim skin)} \geq \Theta   
\end{equation}
where $\Theta$ is a threshold determining the ratio of true and false positives. We empirically determined $\Theta=5$
and set all histogram cells values above this threshold to $255$ (white) and values below $\Theta$ to $0$ (black), 
resulting in a binary image with segmented hands. Figure \ref{fig:handseg} demonstrates the result of the segmentation.

\subsection{Object Detection}
\begin{figure}[thpb]
       \centering
        \includegraphics[width=0.95\textwidth]{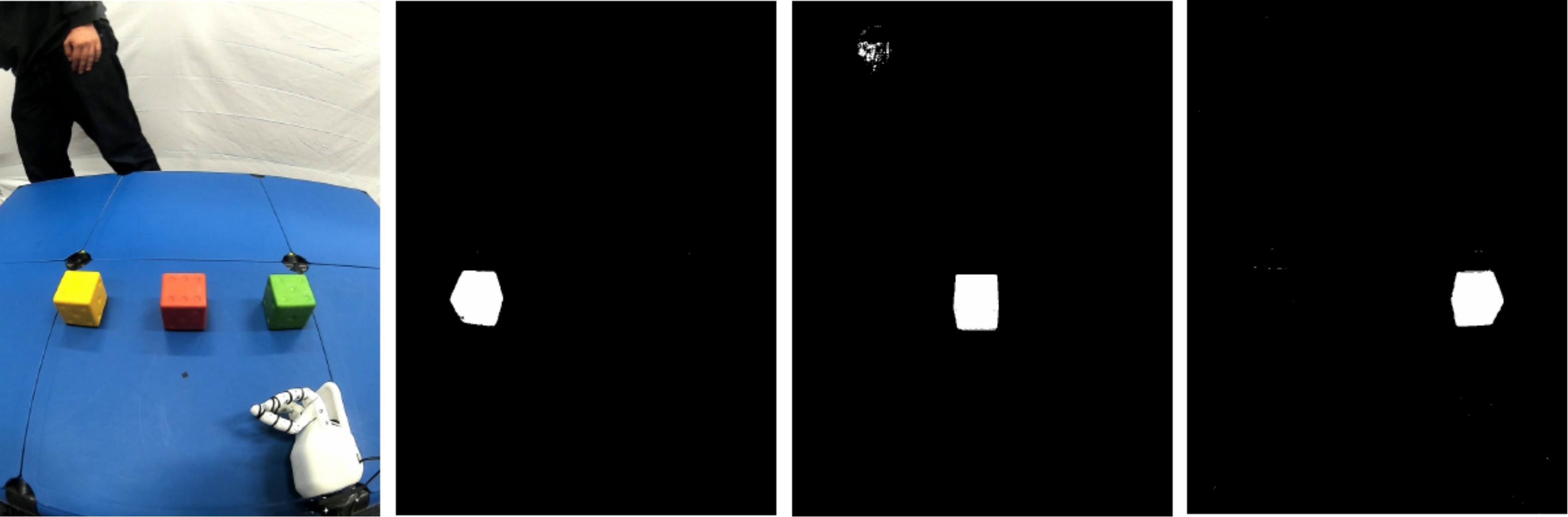}
       \caption{Display and the resultant object segmentation for the yellow, red, and green cube used in our scenario.}
       \label{fig:hsv}
\end{figure}

Our work focuses on the detection and correct recognition of pointing gestures even in cases of object ambiguities, therefore we simplified the objects in the scene to be easily distinguishable by their color. To exploit this object feature, we mapped the RGB images into the HSV color space (Hue, Saturation, Value), which allows a clear color separation using the hue channel. Again, we need to obtain binary masks segmenting the objects from the rest of the image scene, hence after determining the color intervals in the HSV color space, pixels belonging to the color interval are assigned the value $255$, else $0$. The result of the procedure is shown in Figure \ref{fig:hsv}.   

\subsection{Pointing Intention Model}

A major challenge in every gesture scenario is to distinguish a meaningful gesture from arbitrary hand movements. In our scenario, we defined the pointing gesture to be performed with the index finger pointing to an object, which we identified as the most natural behaviour. We use this constraint on the hand pose to model a \textit{pointing intention} of the human subject using a convex hull approach. In brief, a hand can be represented by a finite set of points characterizing the respective hand contour. Computing the convex hull of this contour allows a compact hand representation, where further significant hand features can be obtained. In this work, we used the convex hull algorithm introduced by Jack Sklansky \cite{Sklan_82}, also available in the \textit{OpenCV} library.

\begin{figure}[thpb]
       \centering
        \includegraphics[width=0.9\textwidth]{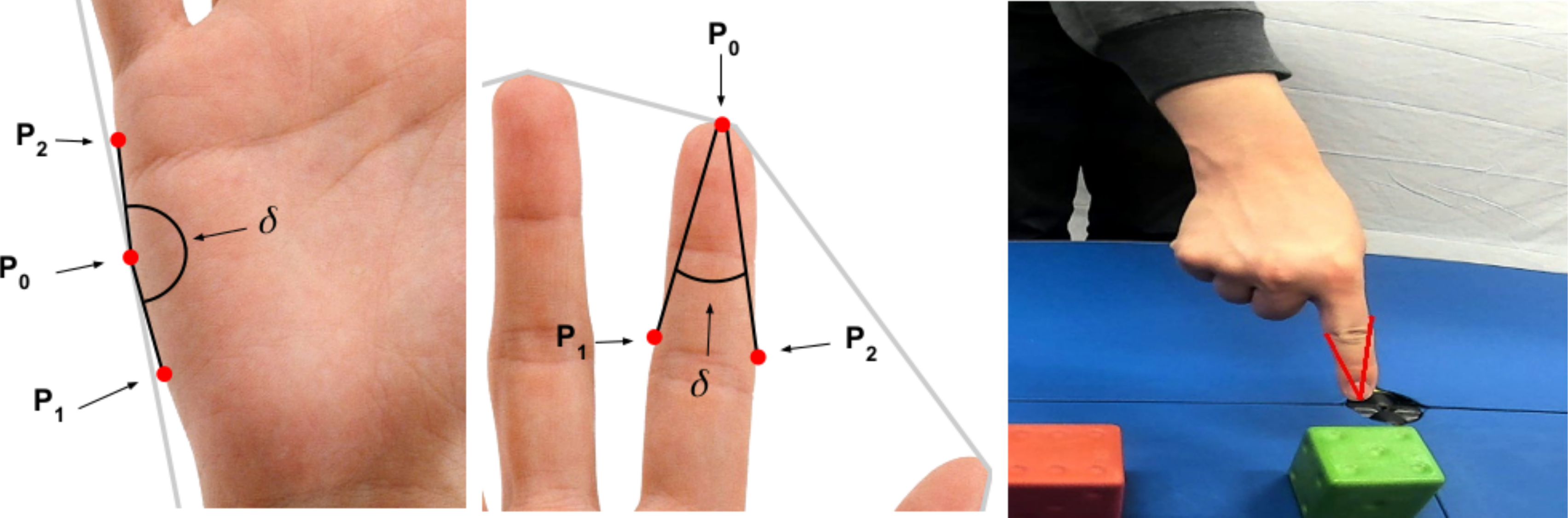}
       \caption{Computation of the pointing characteristic. From points of the convex hull we can compute an angle $\delta$, which is acute when pointing, as shown in the right image.}
       \label{fig:convexHull}
\end{figure}

In addition to the computation of the convex hull, we also consider the so-called convexity defects helping us to determined whether or not a pointing is intended. Based on our pointing gesture definition, we are specifically looking for the fingertips of extended fingers. Thus, the following process is divided into fingertip detection and the verification that exactly one finger is stretched out. Obviously, the fingertip is always part of the convex hull since the robot and the human subject are facing each other, and the camera angle is in a fixed position. 

After the calculation of the convex hull, respectively, the convexity defects, we filtered the latter to avoid disguised local small contour segments.
To compute the correct finger points, we have to take into consideration that the convex hull is implemented as a set of indices mapped to their corresponding points in a contour array. Therefore, for each point $P_0$, two neighboring points $P_1$ and $P_2$ can be computed, resulting in a triangle as depicted in Figure \ref{fig:convexHull}. From the triangle, an angle $\delta$ can be easily obtained. A fingertip is detected, whenever $\delta$ is an acute angle.

\begin{figure}[thpb]
       \centering
        \includegraphics[width=0.8\textwidth]{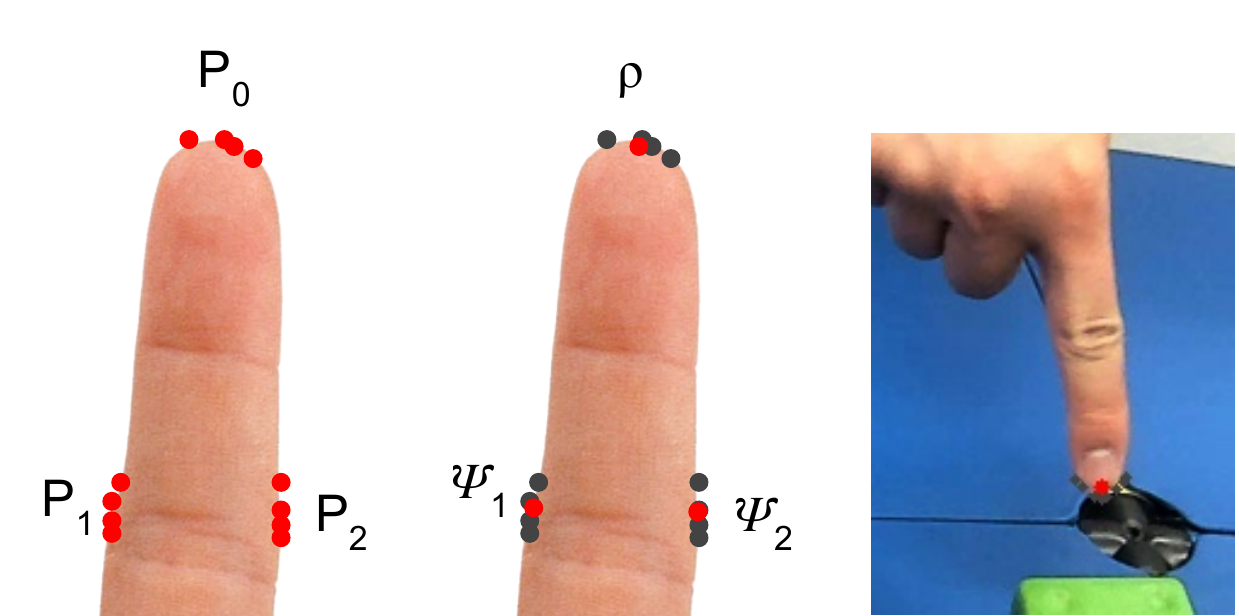}
       \caption{Candidates obeying the angle condition for fingertip detection.}
       \label{fig:fingerCandidates}
\end{figure}
However, the method produces multiple hull points and thus we need to find a mechanism to cluster those points for the final fingertip model. We decided to use hierarchical clustering with single linkage as it creates clusters in a bottom-up fashion without requiring the exact number of clusters. After applying the procedure, we calculate the cluster centroids denoted by $\Psi_1$ and $\Psi_2$ for the corresponding points $P_1$ and $P_2$ and the final fingertip $\rho$, all shown in Figure \ref{fig:fingerCandidates}.

\subsection{The Pointing Array}

After the hand detection, segmentation and the recognition of a pointing intention, we now have to couple the computational steps with the according objects in the scene. The task is whether or not a human subject points to an object, whereby the level of the correct recognition is made more difficult by increasing levels of object ambiguities. 
We first explain the computation of a pointing array for a simple pointing-object association, then we motivate and introduce the usage of the Growing-When-Required network (GWR). 

\begin{figure}[thpb]
       \centering
        \includegraphics[width=0.7\textwidth]{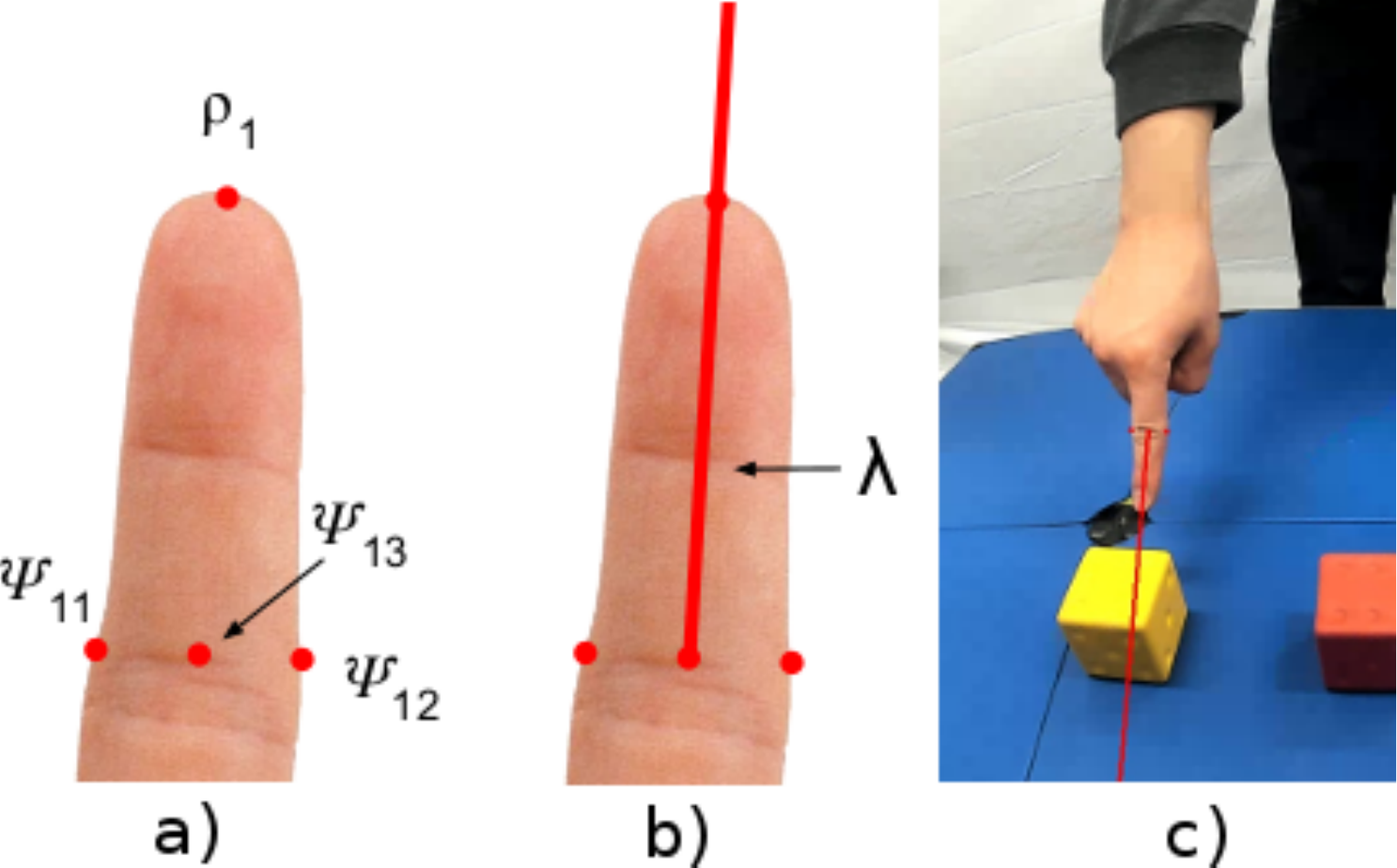}
       \caption{Final computation of the pointing array, following the intuition to extrapolate the direction of the pointing finger. This line is then used to calculate its overlap with an object.}
       \label{fig:point_array}
\end{figure}

In the previous section, we described how to detect the pointing direction with the index finger stretched out. A simple, yet intuitive approach built on top of the previously described method is to extrapolate the direction of the fingertip and calculate whether this pointing array hits an object. Based on the final points obtained from the pointing intention scheme, we calculate another point $\Psi_{13}$ which lies between the two outer finger points $\Psi_{11}$ and $\Psi_{12}$. We connect $\Psi_{13}$ with the fingertip $\rho_1$, as shown in Figure \ref{fig:point_array}a) and extrapolate this line segment called $\lambda$ as demonstrated in Figure \ref{fig:point_array}b). The resultant pointing array hits an object as demonstrated in Figure \ref{fig:point_array}c).

\begin{figure}[thpb]
       \centering
       \includegraphics[width=0.45\textwidth]{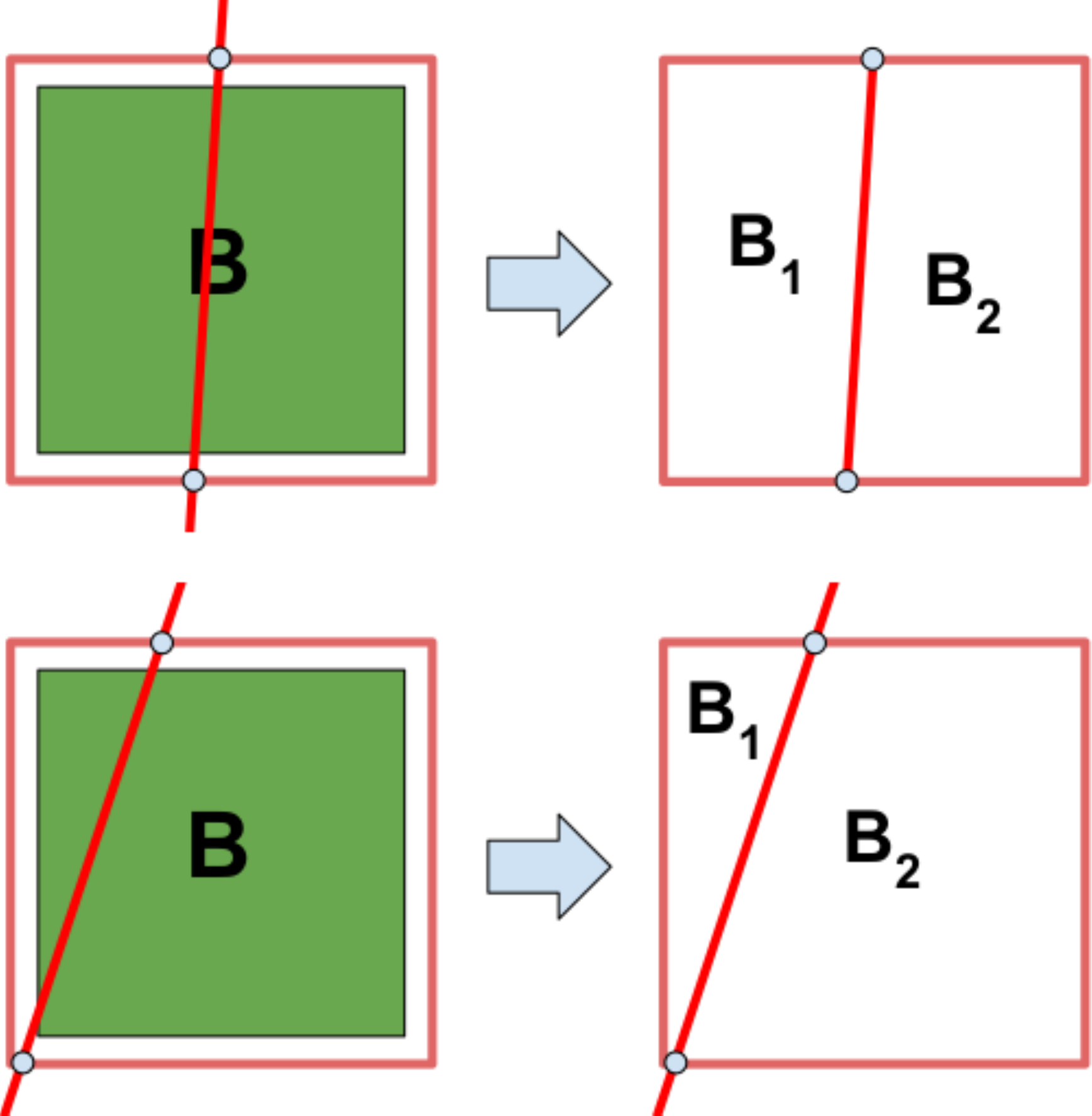}
       \caption{Evaluation of the pointing quality. If the pointing array is clearly targeting the object, as shown in the upper row, the quality measure of this hit is close to 1. More lateral interaction results in less overlap, as demonstrated in the lower row, resulting in a lower value.}
       \label{fig:boxHit}
\end{figure}

We use this concept to finally assess the pointing-object association in our scenario. We start by first detecting all objects present in the scene. Then, we determine which of the objects are hit by the pointing array, i.e. if the array intersects with the bounding box of an object. If an intersection occurred, we measure how accurately the pointing to an object was. For this purpose, we define a quality measure $\phi \in [0;1]$ illustrated in Figure \ref{fig:boxHit}. The object bounding box is divided into $B_1$ and $B_2$ and both areas are calculated. Clearly, if the pointing array hits the bounding box almost centrally, the ratio of both areas is almost identical and thus close to $1$. In contrast, if the pointing is rather lateral, the ratio between the smaller and the bigger areas, as shown in the lower picture of Figure \ref{fig:boxHit}, yields a quality value close to $0$. In this case, the pointing can be assumed to be arbitrary or incorrect. Therefore, we cut off a hit at $\phi<0.2$. In case multiple objects are present, the object yielding the highest value for $\phi$ is considered as an intended hit.

Although the approach is easily implemented, it also has some serious drawbacks negatively impacting the gesture scenario. One issue relates to the extrapolation itself, where confusion of the correct hit is high when more objects are put on the table, e.g. when introducing, from the robot's perspective, an additional row above the existing ones. In line with this, increasing the number of objects also means that more objects have to be detected and tracked, which increases the computation time in cluttered or uncontrolled environments. Finally, our system has so far always assumed an actual pointing by computing an array, even in cases of unintentional gestures.   

\section{Pointing Gestures with Growing-When-Required Networks}

Growing-when-required networks (GWR) \cite{Marsl_02} are a natural extension of Kohonen maps \cite{Kohon_90} by relaxing the assumption of a specific network topology. Still, the underlying idea of unsupervised learning providing a mapping between an input vector space to another, usually quantized, vector space remains. Here, we briefly familiarize the reader with critical computations for GWRs to ensure proper understanding of our computational scheme and evaluation procedure as well as the network parametrization. 

A GWR is a graph composed of a set of vertices $V$, usually referred to in this context as neurons represented as initially random weight vectors, connected by edges $E \subseteq V\times V$. The weights undergo iterative training where the input is presented to the network and, based on some metrics, the node with the most similar associated weight vector is identified as the so-called \textit{best matching unit} (BMU) and, additionally, also a \textit{second best matching unit} (sBMU) whose weights get updated proportionally to the input vector. The principle of strengthening connections between synchronously firing neurons is known as Hebbian learning; due to the exclusive determination of specific neurons this notion is referred to as competitive Hebbian learning.

The crucial difference in the training process compared to Kohonen maps is that for GWRs also the edges play an important role, which is actually the reason why GWRs are a more flexible method to capture a distinct distribution. Every edge in the initial network is zero, while for each iteration the \textit{age} of nodes connected to the BMU is incremented by 1. The connection between nodes is lost when a threshold \textit{$age_{max}$} is encountered, and a node is completely removed when it has no links left to any node. A firing counter for each node manages the network growth. Is a new node identified, the firing counter is 1, and its activity is exponentially decreasing. 

We will now briefly sketch the weight update routines and the parameter used in the training process.
Let $o_i$ be an observation from a complete dataset and the distribution be $p(O)$, i.e. $o_i \in O$. A weight vector is defined as $\omega_n$ for a node $n$. A GWR is initialized with two nodes sampled from $p(O)$. At each iteration, an observation $o_i$ is presented to the network to determine the BMU and sBMU using the Euclidean distance and a link added to the set of edges $E=E \cup \{BMU, sBMU\}$. For existing connections, the $age$ is reset to 0, otherwise. The goodness of the match between the presented input and the resultant BMU is measured by an activity:

\begin{equation}
    a=exp(-||o_i - \omega_{b}||)
\end{equation}
where $||\cdot||$ denotes the Euclidean distance and we use $b$ as the index denoting the BMU for readability reason. The activity is close to 1 for a good fit and close to 0 else. 
A threshold parameter $a_{\theta}$ is used to ensure proper associations between the input space and the resultant representation, thus has to be satisfied before proceeding with the weight updates. This process is controlled by the two learning rates $\eta_b$ and $\eta_n \in [0;1]$, $0<\eta_n<\eta_{b}<1$ and computed as follows:

\begin{eqnarray}
 \label{eq:update_bmu}
    \Delta \omega_{b}=\eta_{b} \times h_{b} \times (o_i - \omega_{b}) \\ 
   \label{eq:update_n}
    \Delta \omega_n=\eta_n \times h_n \times (o_i - \omega_{n}) 
\end{eqnarray}
Similarly, the mentioned firing counter needs an update, denoted by $h_{b}$ and $h_n$:

\begin{eqnarray}
    h_{b}(t)=h_0-\frac{S(t)}{a_{b}}(1-e^{-a_{b}t/\tau_{b}}) \\
    h_n(t)=h_0-\frac{S(t)}{a_{n}}(1-e^{-a_{n}t/\tau_{n}})
\end{eqnarray}
where $h_0$ is the initial firing value and $S(t)$ is the stimulus strength, both set to 1. The variable $t$ is simply an iteration step. The variables $a_b$, $a_n$, $\tau_b$ and $\tau_n$ are constants \cite{Marsl_02}. The distinction between the firing counter of the BMU and the ones for the neighbours is in analogy with the Kohonen map, where surrounding nodes are less updated proportional to the influence of the neighbourhood function. 
After the updates, the \textit{age} is incremented and all edges $e_i>age_{max}$ are deleted.
Finally, the activation and firing counter are also used to discover insufficiently trained nodes or 
incomplete coverage of the input space. For low activity, the update rules in equations \ref{eq:update_bmu} and \ref{eq:update_n} apply. In case of a scarce representation, another threshold for the firing counter $h_{\theta}$ can be used and, in case a certain value is below $h_{\theta}$, a new node is added: $N=N\cup{n\_new}$ with weight vector $\omega_{n\_new}=(\omega_b + o_i)/2$, and edges are created between the BMU and the sBMU.

\subsection{Pointing Using Growing-When-Required Networks}
The input to the GWR should contain unique and meaningful hand properties, thus we decided to use the already identified hand centroid and fingertip positions from the preprocessing procedures explained above. In addition, we integrate the angle of the hand pointing direction. During the performance of different pointing gestures to an object, this angle continuously changes and is, therefore, a complementing feature which might help to increase support for a correct correspondence between the gesture and the object. 

\begin{figure}[thpb]
       \centering
        \includegraphics[width=0.95\textwidth]{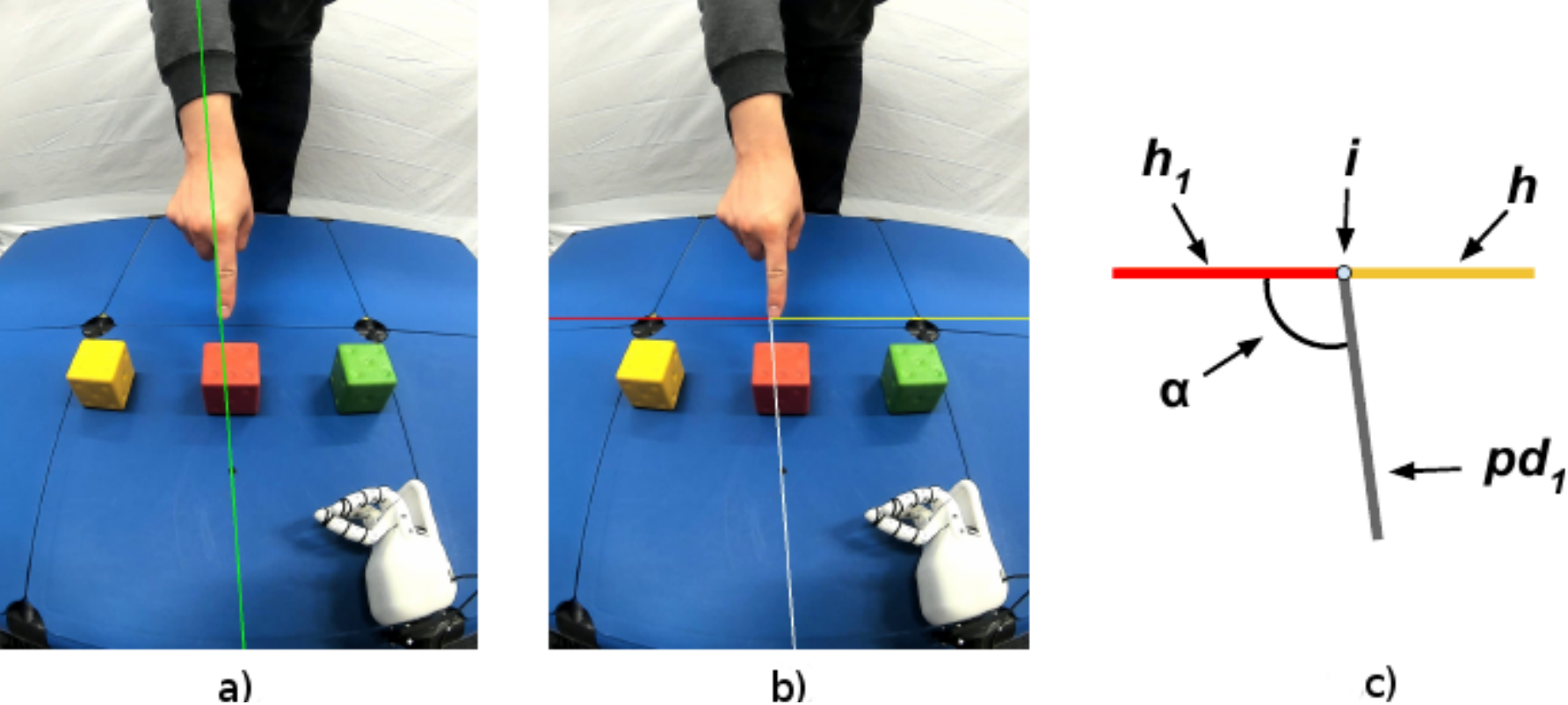}
       \caption{Computation of the angle $\alpha$ signifying the pointing direction by introducing an auxiliary horizontal line intersecting with the pointing direction.}
        \label{fig:gwr_alpha}
\end{figure}

To obtain the angle, we considered the whole contour as the three points extracted for the pointing array are error-prone to noise and, consequently, may introduce prediction errors. 
To approximate the pointing direction given the entire contour we employed the weighted-least-squared error between the pointing line and the contour points. We selected the line with the lowest error and refer to it as $pd$, i.e. pointing direction (\ref{fig:gwr_alpha}a).
We drew a horizontal line $h$ at the level of the fingertip along the whole image width and calculate the intersection with $pd$, which results in two line segments $h_1$ and $pd_1$ (Figure \ref{fig:gwr_alpha}). The specific point is simply called $i$, as shown in Figure \ref{fig:gwr_alpha}c). It is then easy to obtain the angle $\alpha$.

For each frame, the following vector is then established which characterizes the hand shape:

\begin{equation}
V_i=(\alpha_i, c_{xi}, c_{yi}, \rho_{xi}, \rho_{yi})    
\end{equation}
where $\alpha_i$ is the described pointing direction, $c_{*}$ the centroids in the x and y-plane, and $\rho_*$ the respective fingertip coordinates.
Also, a bounding box representing the target of each individual gesture is computed as:
\begin{equation}
B_i=(x_{1i}, y_{1i}, x_{2i}, y_{2i})    
\end{equation}
where $x_{1i}, y_{1i}$ is the top-left coordinate and $x_{2i}, y_{2i}$ the bottom right corner.

\subsection{Definition of the Label Function}
The GWR training is performed in an unsupervised manner and to evaluate the correspondence between the original data and the learnt representation is, in essence, a comparison between vector distances.
However, for realistic implementations, Parisi et al. \cite{Paris_15} showed that GWRs equipped with a label function allow the use of unsupervised learning also for typical classification tasks. In summary, the authors \cite{Paris_15} introduced a hierarchical processing pipeline of feature vectors for action recognition, where one labeling function
was used in the training and a second for the testing phase, i.e. labeling an unseen action.
The fundamental difference between this work and our approach is that we operate in a continuous space (i.e. the table locations) in contrast to discrete action classes.
Therefore, we will explain how we defined our label function.

It is necessary to find a reasonable mapping between a label and the weight changes during the GWR training. We exploit the fact that both weights and labels are continuous points in an $n$-dimensional space. Note, that the weights are 5-dimensional vectors representing our training data, while the label is an annotated bounding box expressed as a 2-dimensional position vector. So, we can apply methods to alter the weights in correspondence with the vector representing the target position. Specifically, this means that changes in the weight consequently change the label, mapping their topology. A correct mapping assumed, similar pointing positions resulting from the pointing gestures are represented by node labels which are spatially close in the network, following the basic intuition of GWRs.

For the concrete implementation of this procedure, we distinguish between the node creation and the node training. Whenever a node is created, we calculate both the mean average between an observation and the BMU weight and, additionally, the mean average of both labels. The resulting value is then assigned to the node. Following the representation of the label, i.e. the position of a bounding box, we define the mean of two of those bounding boxes:

\begin{equation}
    lc_r=(lc_{b}+lc_{o_i})/2
\end{equation}
where $lc_r$ is the centroid of the new node label, $lc_{b}$ denotes the centroid of the BMU label and $lc_{o_i}$ is the centroid obtained from the current observation. Given this measure, a new bounding box is created with the corresponding width and height.

We also update the labels according to:

\begin{equation}
    \Delta lc_{b}=\eta_b \times h_{b} \times (lc_{o_i}-lc_{b}),
\end{equation}
\begin{equation}
    \Delta lc_n=\eta_n \times h_n \times (lc_{o_i}-lc_n)
\end{equation}
where $lc_{b}$ is the BMU centroid label and $lc_n$ the labels of the neighbours as well as the label of the current observation $lc_{o_i}$. Is a new node added, the mean height and width of a bounding box are determined. The update strategy follows the same intuition as the ones for the weights, as sufficiently trained nodes will not significantly change their associated labels.

\subsection{Integration into the Prediction Process}

For easy prediction tasks with no or only a low ambiguity level as implemented for classes $a_1$ and $a_2$, the GWR trained on our recordings returns the label of the BMU. Along with the result, we compare the activation level with a noise threshold to exclude faulty pointings, i.e. unintentional gestures and those diverging significantly from the scenario.
We configured this threshold to $0.5$, because testing incorrect gestures yielded activations around 0.1-0.35, and we want to provide a certain buffer to remain flexible capturing other improper gestures.

\begin{figure}[thpb] 
       \centering
        \includegraphics[width=\textwidth]{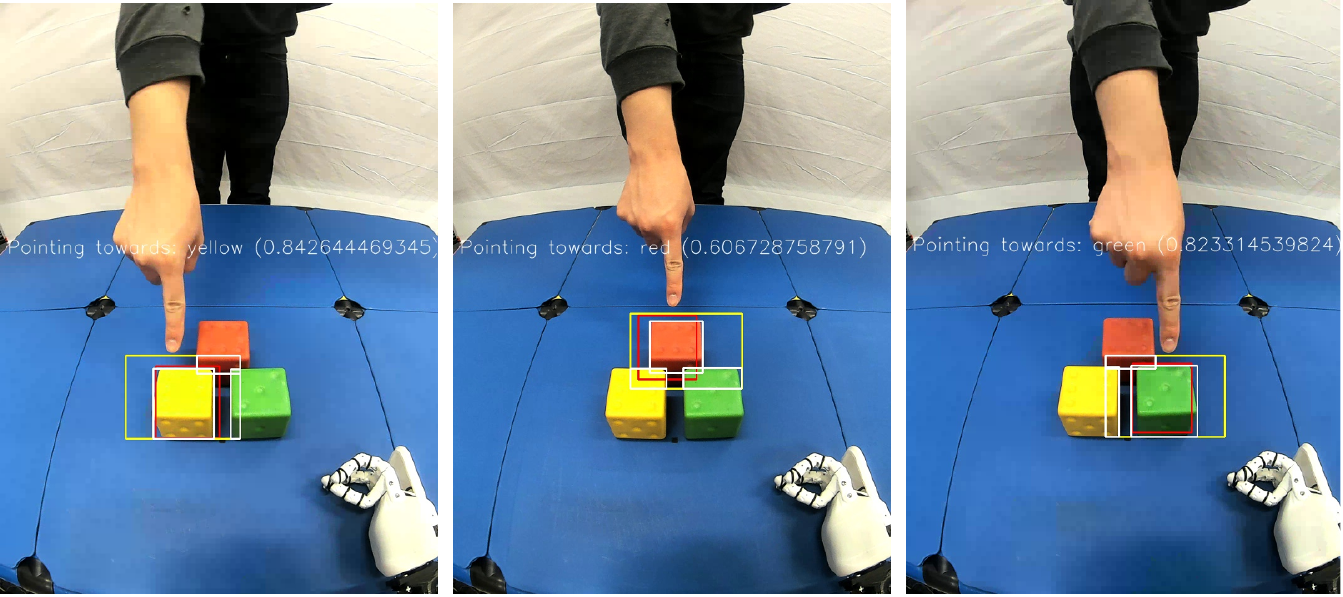}
       \caption{Prediction performance of the GWR when ambiguity is present. The yellow frame is the united area $A$, the white frame is the detected object and the red frame results from processing of the BMU returned by the GWR.}
        \label{fig:pointing_all}
\end{figure}

In case ambiguity is present, also more than one label is returned. From the set of returned BMU labels, we compute the union which we call $A$ representing the positions seen to be most similar to the target direction. From the positions united in A, we calculate the corner points as demonstrated in Figure \ref{fig:gwr_iou}. We set up a function $F_A$ to detect the correct number of labels: 

\begin{figure}[thpb]
       \centering
        \includegraphics[width=0.70\textwidth]{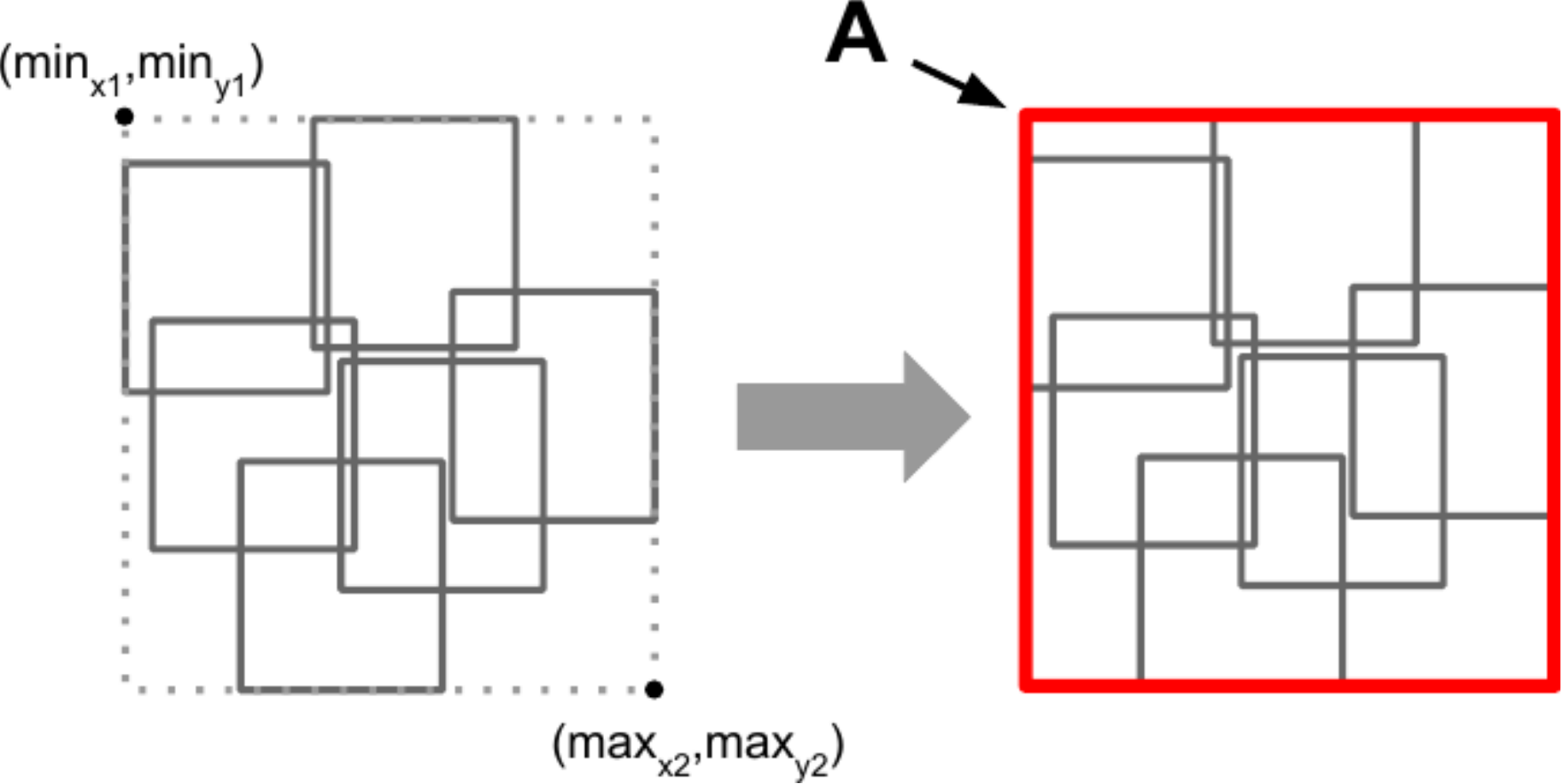}
       \caption{Computation of the final area $A$ after detecting multiple labels returned as the positions of the bounding boxes.}
        \label{fig:gwr_iou}
\end{figure}

\begin{equation}
    F_{A}(\#nodes)=\lceil \#nodes * 0.01+5\rceil
\end{equation}
where $\#nodes$ are the number of nodes in the GWR and the values empirically determined constants. The reason behind is to keep the size of $A$ relatively constant, as the network growth corresponds to node generation, thus the network becomes dense and positions tend to overlap or, at least, show less variance capturing pointing positions in the scene. 
Some results of pointing in ambiguity cases are shown in Figure \ref{fig:pointing_all}, where the numeric values denote the IoU.

\section{Results and Evaluations} 
In the following, we evaluate our pointing scenario obtained from both the computer vision approach and prediction with a GWR network. We also describe the ambiguity detection. 
\subsection{The Computer Vision Approach} 
Table \ref{tab:cv_tps} lists the metrics used to evaluate the success of our system in terms of true positives (TP), false positives (FP) and false negatives (FN). A TP translates to a correctly returned object the subject was targeting at, FP denotes that an object was returned which was not pointed to and an FN means that no object was returned. From those measures, we also compute precision, recall, and the F1 score, which are shown in Table \ref{tab:cv_metrics}. We count a pointing gesture to an object as \textit{correct} or a hit if the actual and predicted object positions share an $IoU\geq 0.5$. Otherwise, we declare it as a \textit{miss}.
Our results demonstrate that our system achieved high precision with an overall value of 96.8\%.
This means that it is likely to get a correct target object, especially underpinned by the 100\% precision value for ambiguity classes $a_1$ to $a_3$. Similarly, both the recall and the F1 score show high values, ranging from 95.87\% to 99.21\% for these classes.

A different picture becomes evident for class $a_4$: the number of misses increases from 7.43\%
to 22.96\%, which consequently affects the recall expressed in a decrease to 71.04\%.
Also, the number of FPs drastically increased from 6 FPs in class $a_3$ to 213, with a direct negative effect on the precision, which is with 92.24\% the lowest value among the classes. 

We identified two factors to explain our obtained results.
First, the proximity of objects is very high, causing variations in the pointing array to be classified as a hit and thus increasing the number of FPs accordingly.
The second factor can be explained by a slight yellow cast in the data recordings we detected in the postprocessing of the data for class $a_4$, which negatively affects the hand detection as it relies on a learnt color distribution. Thus, the segmentation is more likely to fail for those frames and, additionally, influence the pointing array to deviate strongly from the intended pointing direction. Therefore, the intersection of the pointing array and the target object is less likely to occur.


\begin{table}
\caption{Metrics for the computer vision approach}
\label{tab:cv_tps}
\begin{tabular}{llllll}
\hline\noalign{\smallskip}
 &  TP   & FP  & FN   & Miss & Total \\
\noalign{\smallskip}\hline\noalign{\smallskip}
$a_1$                  & 1248 & 0   & 20   & 20   & 1268  \\ 
$a_2$                  & 1428 & 4   & 119  & 115  & 1547  \\ 
$a_3$                  & 1356 & 6   & 55   & 49   & 1411  \\ 
$a_4$                  & 2414 & 203 & 984  & 781  & 3398  \\ 
Total                  & 6446 & 213 & 1178 & 965  & 7624  \\ 
\noalign{\smallskip}\hline
\end{tabular}
\end{table}


\begin{table}
\caption{Evaluation of the computer vision approach}
\label{tab:cv_metrics}
\begin{tabular}{llllll}
\hline\noalign{\smallskip}
 & Precision & Recall & F1 & Miss \\ 
\noalign{\smallskip}\hline\noalign{\smallskip}
{$a_1$} & 100.0 & 98.42 & 99.21 & 1.58 \\
{$a_2$} & 99.72 & 92.31 & 95.87 & 7.43 \\
{$a_3$} & 99.56 & 96.1 & 97.8 & 3.47 \\ 
{$a_4$} & 92.24 & 71.04 & 80.27 & 22.96 \\ 
{Total} & 96.8 & 84.55 & 90.26 & 12.65 \\
\noalign{\smallskip}\hline
\end{tabular}
\end{table}

To summarize, our baseline approach based on computer vision techniques shows astonishingly good performance for the first three ambiguity classes $a_1$ to $a_3$. We conclude that for these classes, it is worth to look into procedures which are easily implemented and, as we demonstrate, robust in terms of the mapping between the pointing gesture and the targeted object. However, increasing the object ambiguity as shown for class $a_4$ rapidly deteriorates the performance of our interface. Therefore, we propose a flexible modelling approach using unsupervised learning, specifically, involving a specific implementation of a growing-when-required network (GWR), which we explain in the next section.

\subsection{The GWR Network Model} 

The parameter $a_T$ denotes the insertion threshold, which has been identified to be the most crucial GWR parameter \cite{Marsl_02} because it controls the network growth. Following related work \cite{Paris_15}, we set $a_T\in\{0.85, 0.90, 0.95\}$ and the number of epochs: ${30, 50, 100}$.
For training the GWR, we used 3-fold cross-validation.
Table \ref{tab:gwr_param} depicts the GWR network parameters which are fixed throughout the training. 


\begin{table}
\caption{GWR Network Parameters}
\label{tab:gwr_param}
\begin{tabular}{llllllllll}
\hline\noalign{\smallskip}
$e_b$ & $e_n$ & $h_T$ & $\tau_b$ & $\tau_n$ & $a_b$ & $a_n$ & $h_0$ & $age_{max}$ & $nb_{max}$ \\
\noalign{\smallskip}\hline\noalign{\smallskip}
0.1 & 0.01 & 0.1 & 0.3 & 0.1 & 1.05 & 1.05 & 1.0 & 200 & 6 \\ 
\noalign{\smallskip}\hline
\end{tabular}
\end{table}

Here, $e_b$ is the learning rate for the best matching unit (BMU), $e_n$ is the learning
rate for the BMU neighbors, $h_T$ is the firing threshold, and $\tau_b$, $\tau_n$, $a_b$, $a_n$, $h_0$ are the
habituation parameter for the BMU and its neighbors. The parameters $age_{max}$ and
$nb_{max}$ specify the maximum age of an edge before deletion and the maximum
number of neighbors a node can have. 

\begin{table}
\caption{Exemplary evaluation of the GWR network size}
\label{tab:gwr_growth}
\begin{tabular}{lllll}
\hline\noalign{\smallskip}
$a_T$ & epochs & \#nodes & \#edges & error \\
\noalign{\smallskip}\hline\noalign{\smallskip}
 & 30 & 263 & 633 & 0.0597 \\
0.85 & 50 & 332 & 861 & 0.0557 \\
 & 100 & 448 & 1217 & 0.0522 \\
 & 30 & 541 & 1422 & 0.0461 \\
0.90 & 50 & 677 & 1797 & 0.0422 \\
    & 100 & 890 & 2409 & 0.0394 \\
    & 30 & 1543 & 4206 & 0.0268 \\
0.95 & 50 & 1930 & 5222 & 0.0261 \\
    & 100 & 2606 & 7263 & 0.0250 \\
\noalign{\smallskip}\hline
\end{tabular}
\end{table}

Table \ref{tab:gwr_growth} shows the averaged result from the cross-validation in terms of the network size and the error, the latter denoting the overall distances between the BMU returned by the network and the actual target position. As expected, the higher the value of $a_T$ is, the more nodes and respective edges are created. Consequently, we obtained the lowest error for $a_T=0.95$ but nevertheless we cannot be sure that nodes are redundant and, more importantly, we may trap into overfitting as the GWR covers the learnt input space but may fail to generalize to unseen positions. 
Therefore, we conducted a detailed evaluation on i.) the performance of a GWR w.r.t. pointing positions, unveiling the potentially best candidate network balancing network growth and performance, denoted as `Individual GWR' and ii.) the integration of the object(s) and their respective ambiguity, denoted as `GWR Performance with Objects'.

\subsection{GWR Pointing Predictions}
The results reported in tables \ref{tab:gwr_a85}, \ref{tab:gwr_a90}, and \ref{tab:gwr_a95} are the mean and standard deviation averaged over the three folds using the test set. We report precision, recall, the F1 score, as well as the number of misses in \%.
\begin{table}
\caption{Individual GWR with $a_T=0.85$}
\label{tab:gwr_a85}
\begin{tabular}{lllll}
\noalign{\smallskip}\hline\noalign{\smallskip}
       & Precision (\%) & Recall (\%) & F1 (\%)  & Misses (\%) \\
\hline\noalign{\smallskip}
       & 30 epochs & IoU:0.677 $\pm$ 0.004 & & \\
\noalign{\smallskip}\hline\noalign{\smallskip}
  {$a_1$}  &  {100.00 $\pm$ 0.0} &   {74.43 $\pm$ 2.2} &   {85.32 $\pm$ 1.4} &   {25.57 $\pm$ 2.2} \\ {$a_2$}  &   {100.00 $\pm$ 0.0} &   {87.23 $\pm$ 1.5} &   {93.17 $\pm$ 0.8} &   {12.77 $\pm$ 1.5} \\ {$a_3$}  &   {99.76 $\pm$ 0.0}  &   {88.6 $\pm$ 3.1}  &   {93.82 $\pm$ 1.8} &   {11.18 $\pm$ 3.1} \\ {$a_4$}  &   {100.00 $\pm$ 0.0} &   {90.60 $\pm$ 1.0} &   {95.11 $\pm$ 0.6} &   {9.31 $\pm$ 1.0}  \\ {$\sum$} &   {99.95 $\pm$ 0.0}  &   {86.89 $\pm$ 0.2} &   {92.97 $\pm$ 0.1} &   {13.07 $\pm$ 0.2} \\
\noalign{\smallskip}\hline
& 50 epochs & IoU:0.685 $\pm$ 0.005 & & \\
\hline\noalign{\smallskip}
$a_1$                        & 100.00 $\pm$ 0.0                      & 80.73 $\pm$ 8.0                      & 89.12 $\pm$ 4.9                      & 19.27 $\pm$ 8.0                      \\ 
$a_2$                        & 100.00 $\pm$ 0.0                      & 89.2 $\pm$ 1.6                       & 94.28 $\pm$ 0.9                      & 10.8 $\pm$ 1.6                       \\ \
$a_3$                        & 99.92 $\pm$ 0.1                       & 84.26 $\pm$ 3.8                      & 91.38 $\pm$ 2.3                      & 15.68 $\pm$ 3.8                      \\ 
$a_4$                        & 100.00 $\pm$ 0.0                      & 90.82 $\pm$ 0.5                      & 95.19 $\pm$ 0.3                      & 9.18 $\pm$ 0.5                       \\ 
$\sum$                       & 99.98 $\pm$ 0.0                       & 87.62 $\pm$ 2.3                      & 93.38 $\pm$ 1.3                      & 12.37 $\pm$ 2.3                      \\ 
\hline\noalign{\smallskip}
& 100 epochs & IoU:0.713 $\pm$ 0.001 & & \\
\hline\noalign{\smallskip}
$a_1$                        & 100.00 $\pm$ 0.0                      & 85.12 $\pm$ 5.2                      & 91.88 $\pm$ 3.0                      & 14.88 $\pm$ 5.2                      \\ 
$a_2$                        & 100.00 $\pm$ 0.0                      & 90.87 $\pm$ 1.0                      & 95.21 $\pm$ 0.5                      & 9.13 $\pm$ 1.0                       \\ 
$a_3$                        & 99.91 $\pm$ 0.1                       & 87.57 $\pm$ 1.3                      & 93.33 $\pm$ 0.7                      & 12.35 $\pm$ 1.4                      \\ 
$a_4$                        & 100.00 $\pm$ 0.0                      & 94.23 $\pm$ 1.8                      & 97.02 $\pm$ 1.0                      & 5.77 $\pm$ 1.8                       \\ 
$\sum$                       & 99.99 $\pm$ 0.0                       & 90.79 $\pm$ 1.1                      & 95.17 $\pm$ 0.6                      & 9.19 $\pm$ 1.0                       \\ 
\hline\noalign{\smallskip}
\end{tabular}
\end{table}

\begin{table}
\caption{Individual GWR with $a_T=0.90$}
\label{tab:gwr_a90}
\begin{tabular}{lllll}
\noalign{\smallskip}\hline\noalign{\smallskip}
       & Precision (\%) & Recall (\%) & F1 (\%)  & Misses (\%) \\
\hline\noalign{\smallskip}
       & 30 epochs & IoU:0.725 $\pm$ 0.0 & & \\
\noalign{\smallskip}\hline\noalign{\smallskip}
$a_1$  & 100.0 $\pm$ 0.0 & 91.03 $\pm$ 2.2 & 95.29 $\pm$ 1.2 & 8.97 $\pm$ 2.2 \\
$a_2$  & 100.0 $\pm$ 0.0 & 94.46 $\pm$ 1.1 & 97.15 $\pm$ 0.6 & 5.54 $\pm$ 1.1 \\
$a_3$  & 100.0 $\pm$ 0.0 & 96.67 $\pm$ 0.5 & 98.31 $\pm$ 0.2 & 3.33 $\pm$ 0.5 \\
$a_4$  & 100.0 $\pm$ 0.0 & 95.34 $\pm$ 0.1 & 97.62 $\pm$ 0.0 & 4.66 $\pm$ 0.1 \\
$\sum$ & 100.0 $\pm$ 0.0 & 94.68 $\pm$ 0.2 & 97.27 $\pm$ 0.1 & 5.32 $\pm$ 0.2 \\
\hline\noalign{\smallskip}
       & 50 epochs & IoU:0.736 $\pm$ 0.002 & & \\
\noalign{\smallskip}\hline\noalign{\smallskip}
$a_1$  & 100.0 $\pm$ 0.0 & 92.77 $\pm$ 1.7 & 96.24 $\pm$ 0.9 & 7.23 $\pm$ 1.7 \\ 
$a_2$  & 100.0 $\pm$ 0.0 & 95.37 $\pm$ 1.1 & 97.63 $\pm$ 0.6 & 4.63 $\pm$ 1.1 \\ 
$a_3$  & 100.0 $\pm$ 0.0 & 95.92 $\pm$ 1.1 & 97.92 $\pm$ 0.6 & 4.08 $\pm$ 1.1 \\ 
$a_4$  & 100.0 $\pm$ 0.0 & 97.53 $\pm$ 0.2 & 98.75 $\pm$ 0.1 & 2.47 $\pm$ 0.2 \\ 
$\sum$ & 100.0 $\pm$ 0.0 & 95.99 $\pm$ 0.4 & 97.95 $\pm$ 0.2 & 4.01 $\pm$ 0.4 \\ 
\noalign{\smallskip}\hline\noalign{\smallskip}
       & 100 epochs & IoU:0.76 $\pm$ 0.004 & & \\
\noalign{\smallskip}\hline\noalign{\smallskip}
$a_1$  & 100.0 $\pm$ 0.0 & 93.75 $\pm$ 2.4 & 96.76 $\pm$ 1.3 & 6.25 $\pm$ 2.4 \\ 
$a_2$  & 100.0 $\pm$ 0.0 & 95.08 $\pm$ 2.0 & 97.47 $\pm$ 1.0 & 4.92 $\pm$ 2.0 \\ 
$a_3$  & 100.0 $\pm$ 0.0 & 98.18 $\pm$ 0.7 & 99.08 $\pm$ 0.3 & 1.82 $\pm$ 0.7 \\ 
$a_4$  & 100.0 $\pm$ 0.0 & 97.32 $\pm$ 0.4 & 98.64 $\pm$ 0.2 & 2.68 $\pm$ 0.4 \\ 
$\sum$ & 100.0 $\pm$ 0.0 & 96.42 $\pm$ 0.4 & 98.17 $\pm$ 0.2 & 3.58 $\pm$ 0.4 \\
\noalign{\smallskip}\hline\noalign{\smallskip}
\end{tabular}
\end{table}

\begin{table}
\caption{Individual GWR with $a_T=0.95$}
\label{tab:gwr_a95}
\begin{tabular}{lllll}
\noalign{\smallskip}\hline\noalign{\smallskip}
       & Precision (\%) & Recall (\%) & F1 (\%)  & Misses (\%) \\
\hline\noalign{\smallskip}
       & 30 epochs & IoU:0.808 $\pm$ 0.001 & & \\
\noalign{\smallskip}\hline\noalign{\smallskip}
$a_1$  & 100.0 $\pm$ 0.0 & 97.63 $\pm$ 0.6 & 98.8 $\pm$ 0.3  & 2.37 $\pm$ 0.6 \\ 
$a_2$  & 100.0 $\pm$ 0.0 & 97.8 $\pm$ 0.3  & 98.89 $\pm$ 0.2 & 2.2 $\pm$ 0.3  \\ 
$a_3$  & 100.0 $\pm$ 0.0 & 98.62 \textbackslash{}m 0.3  & 99.31 $\pm$ 0.1 & 1.38 $\pm$ 0.3 \\ 
$a_4$  & 100.0 $\pm$ 0.0 & 99.17 $\pm$ 0.2 & 99.59 $\pm$ 0.1 & 0.83 $\pm$ 0.2 \\ 
$\sum$ & 100.0 $\pm$ 0.0 & 98.55 $\pm$ 0.2 & 99.27 $\pm$ 0.1 & 1.45 $\pm$ 0.2 \\ \noalign{\smallskip}\hline\noalign{\smallskip}
       & 50 epochs & IoU:0.813 $\pm$ 0.001 & & \\
\noalign{\smallskip}\hline\noalign{\smallskip}
$a_1$  & 100.0 $\pm$ 0.0 & 97.03 $\pm$ 1.3 & 98.49 $\pm$ 0.7 & 2.97 $\pm$ 1.3 \\ 
$a_2$  & 100.0 $\pm$ 0.0 & 97.61 $\pm$ 0.4 & 98.79 $\pm$ 0.2 & 2.39 $\pm$ 0.4 \\ 
$a_3$  & 100.0 $\pm$ 0.0 & 98.47 $\pm$ 0.8 & 99.23 $\pm$ 0.4 & 1.53 $\pm$ 0.8 \\ 
$a_4$  & 100.0 $\pm$ 0.0 & 99.19 $\pm$ 0.5 & 99.59 $\pm$ 0.2 & 0.81 $\pm$ 0.5 \\ 
$\sum$ & 100.0 $\pm$ 0.0 & 98.36 $\pm$ 0.3 & 99.17 $\pm$ 0.1 & 1.64 $\pm$ 0.3 \\ 
\noalign{\smallskip}\hline\noalign{\smallskip}
       & 100 epochs & IoU:0.824 $\pm$ 0.001 & & \\
\noalign{\smallskip}\hline\noalign{\smallskip}
$a_1$  & 100.0 $\pm$ 0.0 & 97.24 $\pm$ 0.6 & 98.6 $\pm$ 0.3  & 2.76 $\pm$ 0.6 \\ 
$a_2$  & 100.0 $\pm$ 0.0 & 98.53 $\pm$ 1.0 & 99.26 $\pm$ 0.5 & 1.47 $\pm$ 1.0 \\ 
$a_3$  & 100.0 $\pm$ 0.0 & 99.44 $\pm$ 0.3 & 99.72 $\pm$ 0.2 & 0.56 $\pm$ 0.3 \\ 
$a_4$  & 100.0 $\pm$ 0.0 & 99.38 $\pm$ 0.3 & 99.69 $\pm$ 0.2 & 0.62 $\pm$ 0.3 \\ 
$\sum$ & 100.0 $\pm$ 0.0 & 98.86 $\pm$ 0.3 & 99.43 $\pm$ 0.1 & 1.14 $\pm$ 0.3 \\ 
\noalign{\smallskip}\hline\noalign{\smallskip}
\end{tabular}
\end{table}

A first result to be highlighted for the prediction performance is the high precision over all numbers of epochs and different
values for the parameter $a_T$, accompanied by a standard deviation of nearly 0. This means that, overall, the GWR is almost perfectly able to represent a desired pointing gesture. A more nuanced picture becomes evident for the recall, which differs greatly over the numbers of epochs, best demonstrated for $a_T=0.85$. Also, the standard deviations show many variations, regulated only by higher numbers of epochs and for $a_T=0.95$. As a trade-off between both measures, the F1 score depicts reasonable performance for 30 epochs and superior performance for 50 and 100 epochs over all ambiguity classes for $a_T=0.85$.
As an example, the F1 score for class $a_1$ is 85.32\% for 30 epochs and increases to 91.88\% for 100 epochs. Similarly, the recall improves from 74.43\% to 85.12\%, which can be explained when considering also the relatively high number of \textit{misses}, affecting the recall. Although the training epochs improve on the number, presumably by the higher amount of nodes produced by the GWR, increasing the value for $a_T$ seems to better capture the negative cases.
Finally, for the remaining classes and considering also the mean values among epochs when $a_T=0.85$, the differences are less pronounced. Interestingly, also the IoU does not significantly change.
Thus, the number of training epochs may be lowered when computational resources are limited.

When $a_T=0.9$, improvements are clearly visible for the recall and the number of misses, yielding superior performance expressed in the F1 score when compared to $a_T=0.85$. Here, the difference between epochs is negligible both for the individual evaluation of the ambiguity classes and the mean values. Similar results are obtained for $a_T=0.95$, which, finally, demonstrates the lowest number of misses and highest IoU with small standard deviation. As shown in table \ref{tab:gwr_growth}, increasing $a_T$ leads to a stronger growth of the GWR, but the relationship between network size and performance is not linear. 30 epochs and $a_T=0.85$, i.e. 263 nodes has more impact on the performance compared to $a_T=0.9$ with 541 nodes. However, the influence of the network size w.r.t. the performance vanishes for parameter $a_T=0.9$ and $a_T=0.95$, and there might be a saturation point where only redundant nodes are created. We, therefore, conclude that reasonable performance can already be achieved with a lower number of epochs and smaller network size. 

\subsection{GWR Predictions with Object Detection}
\begin{figure}[h!]
       \centering
       \includegraphics[width=\textwidth]{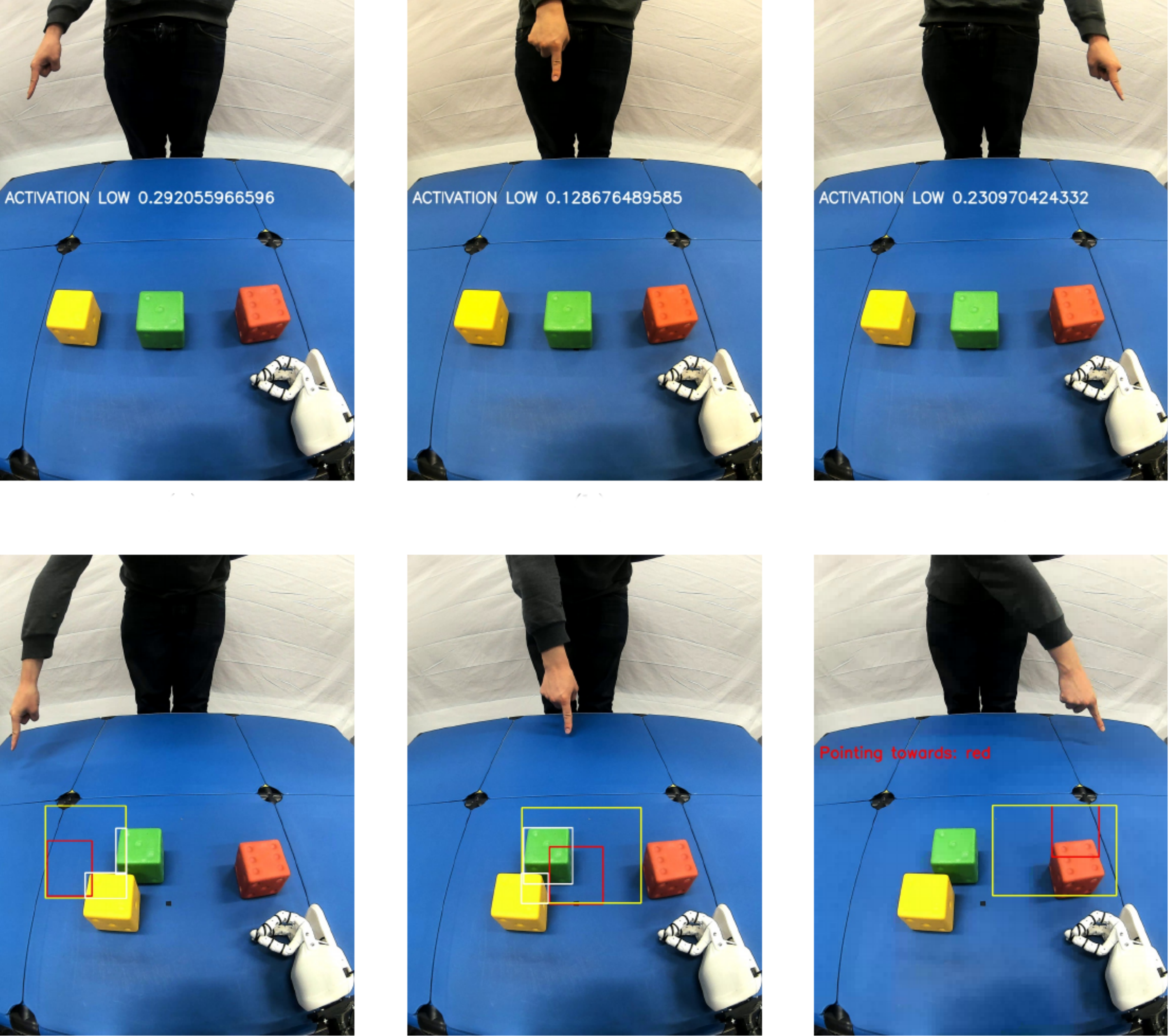}
       \caption{Evaluation of the learnt GWR topology. Upper row: if pointing occurs too distant from the actual setting, the GWR has only low activations and does not classify the pointing. However, the network returns the closest possible spatial location the subject is pointing to with its associated object for positions disjoint from the training set, as demonstrated in the lower row.}
       \label{fig:topo}
\end{figure}

The GWR networks have so far shown good performance and robust classification between the actual and predicted pointing direction to an object. However, we cannot make any statements about correctly returned objects, especially when applying the networks to ambiguous object arrangements. To conclude on the final capabilities of GWR networks in our pointing gesture scenario, we additionally performed experiments integrating the object(s). We used the 40th frame of each individual pointing video, because this frame delivered the information we needed. We extracted the color information from the test frame and compared it with the actual color. In case the colors matched, we counted this result as a TP, otherwise we declared it both a FN and FP. In case no object was detected at all, we counted this as a FN and a miss. This procedure can be applied to the ambiguity class $a_1$ but it did not cover any ambiguous object configurations.
Therefore, we also implemented an ambiguity detection by checking the activation level of the BMU returned by the network: if it was below a predetermined noise threshold $n_T$, we considered the observation as noise, otherwise we approved an existing ambiguity in the current scenario. We then checked the IoU between the predicted and all object positions, followed by the classification into the metrics as just described. Again, we calculated precision, recall, the F1 score and the number of misses, averaged over the 3-fold cross-validation including the standard deviation. Figure \ref{fig:topo} shows exemplary a pointing scenario, demonstrating the activation level for different pointing gestures. The concrete results are listed in tables \ref{tab:gwr_predicta85}, \ref{tab:gwr_predicta90}, and \ref{tab:gwr_predicta95}.

\begin{table}
\caption{GWR Performance with Objects with $a_T=0.85$}
\label{tab:gwr_predicta85}
\begin{tabular}{lllll}
\noalign{\smallskip}\hline\noalign{\smallskip}
       & Precision (\%) & Recall (\%) & F1 (\%)  & Misses (\%) \\
\hline\noalign{\smallskip}
       & 30 epochs & IoU:0.691 $\pm$ 0.003 & & \\
\noalign{\smallskip}\hline\noalign{\smallskip}
$a_1$  & 100.0 $\pm$ 0.0 & 100.0 $\pm$ 0.0 & 100.0 $\pm$ 0.0 & 0.0 $\pm$ 0.0   \\
$a_2$  & 100.0 $\pm$ 0.0 & 94.6 $\pm$ 0.8  & 97.22 $\pm$ 0.4 & 5.4 $\pm$ 0.8   \\
$a_3$  & 98.54 $\pm$ 1.0 & 85.0 $\pm$ 0.5  & 91.27 $\pm$ 0.3 & 13.72 $\pm$ 1.3 \\
$a_4$  & 100.0 $\pm$ 0.0 & 85.95 $\pm$ 1.6 & 92.44 $\pm$ 0.9 & 14.05 $\pm$ 1.6 \\
$\sum$ & 99.75 $\pm$ 0.2 & 89.84 $\pm$ 0.5 & 94.54 $\pm$ 0.3 & 9.93 $\pm$ 0.4  \\
\noalign{\smallskip}\hline\noalign{\smallskip}
       & 50 epochs & IoU:0.7 $\pm$ 0.002 & & \\
\noalign{\smallskip}\hline\noalign{\smallskip}
$a_1$  & 100.0 $\pm$ 0.0 & 100.0 $\pm$ 0.0 & 100.0 $\pm$ 0.0 & 2.97 $\pm$ 1.3  \\ 
$a_2$  & 100.0 $\pm$ 0.0 & 95.39 $\pm$ 0.7 & 97.64 $\pm$ 0.4 & 4.61 $\pm$ 0.7  \\ 
$a_3$  & 99.43 $\pm$ 0.6 & 82.7 $\pm$ 3.8  & 90.26 $\pm$ 2.5 & 16.84 $\pm$ 3.3 \\ 
$a_4$  & 99.97 $\pm$ 0.0 & 85.41 $\pm$ 2.0 & 92.11 $\pm$ 1.2 & 14.56 $\pm$ 2.0 \\ 
$\sum$ & 99.9 $\pm$ 0.1  & 89.37 $\pm$ 0.9 & 94.34 $\pm$ 0.6 & 10.54 $\pm$ 0.8 \\ 
\noalign{\smallskip}\hline\noalign{\smallskip}
       & 100 epochs & IoU:0.715 $\pm$ 0.004 & & \\
\noalign{\smallskip}\hline
$a_1$  & 100.0 $\pm$ 0.0 & 100.0 $\pm$ 0.0 & 100.0 $\pm$ 0.0 & 100.0 $\pm$ 0.0 \\ 
$a_2$  & 100.0 $\pm$ 0.0 & 96.33 $\pm$ 0.3 & 98.13 $\pm$ 0.1 & 3.67 $\pm$ 0.3  \\ 
$a_3$  & 99.85 $\pm$ 0.2 & 85.6 $\pm$ 3.4  & 92.14 $\pm$ 2.0 & 14.27 $\pm$ 3.5 \\ 
$a_4$  & 100.0 $\pm$ 0.0 & 87.92 $\pm$ 3.1 & 93.55 $\pm$ 1.7 & 12.08 $\pm$ 3.1 \\ 
$\sum$ & 99.97 $\pm$ 0.0 & 91.2 $\pm$ 1.6  & 95.38 $\pm$ 0.9 & 8.77 $\pm$ 1.7  \\
\noalign{\smallskip}\hline
\end{tabular}
\end{table}

\begin{table}
\caption{GWR Performance with Objects with $a_T=0.90$}
\label{tab:gwr_predicta90}
\begin{tabular}{lllll}
\noalign{\smallskip}\hline\noalign{\smallskip}
       & Precision (\%) & Recall (\%) & F1 (\%)  & Misses (\%) \\
\hline\noalign{\smallskip}
       & 30 epochs & IoU:0.713 $\pm$ 0.0 & & \\
\noalign{\smallskip}\hline\noalign{\smallskip}
$a_1$  & 100.0 $\pm$ 0.0 & 100.0 $\pm$ 0.0 & 100.0 $\pm$ 0.0 & 0.0 $\pm$ 0.0   \\ 
$a_2$  & 100.0 $\pm$ 0.0 & 98.33 $\pm$ 0.4 & 99.16 $\pm$ 0.2 & 1.67 $\pm$  0.4 \\ 
$a_3$  & 100.0 $\pm$ 0.0 & 91.5 $\pm$ 1.3  & 95.56 $\pm$ 0.7 & 8.5 $\pm$ 1.3   \\ 
$a_4$  & 100.0 $\pm$ 0.0 & 90.77 $\pm$ 1.9 & 95.15 $\pm$ 1.0 & 9.23 $\pm$ 1.9  \\ 
$\sum$ & 100.0 $\pm$ 0.0 & 93.94 $\pm$ 0.6 & 96.88 $\pm$ 0.3 & 6.06 $\pm$ 0.6  \\ 
\noalign{\smallskip}\hline\noalign{\smallskip}
       & 50 epochs & IoU:0.719 $\pm$ 0.002 & & \\
\noalign{\smallskip}\hline\noalign{\smallskip}
$a_1$  & 100.0 $\pm$ 0.0 & 100.0 $\pm$ 0.0 & 100.0 $\pm$ 0.0 & 0.0 $\pm$ 0.0   \\ 
$a_2$  & 100.0 $\pm$ 0.0 & 98.55 $\pm$ 0.5 & 99.27 $\pm$ 0.2 & 1.45 $\pm$ 0.5  \\ 
$a_3$  & 100.0 $\pm$ 0.0 & 89.63 $\pm$ 0.1 & 94.53 $\pm$ 0.1 & 10.37 $\pm$ 0.1 \\ 
$a_4$  & 99.97 $\pm$ 0.0 & 93.16 $\pm$ 0.9 & 96.44 $\pm$ 0.5 & 6.81 $\pm$ 0.9  \\ 
$\sum$ & 99.99 $\pm$ 0.0 & 94.71 $\pm$ 0.3 & 97.28 $\pm$ 0.2 & 5.28 $\pm$ 0.3  \\
\noalign{\smallskip}\hline\noalign{\smallskip}
       & 100 epochs & IoU:0.735 $\pm$ 0.002 & & \\
\noalign{\smallskip}\hline\noalign{\smallskip}
$a_1$  & 100.0 $\pm$ 0.0 & 100.0 $\pm$ 0.0 & 100.0 $\pm$ 0.0 & 0.0 $\pm$ 0.0   \\ 
$a_2$  & 100.0 $\pm$ 0.0 & 98.44 $\pm$ 1.2 & 99.21 $\pm$ 0.6 & 1.56 $\pm$ 1.2  \\ 
$a_3$  & 99.92 $\pm$ 0.1 & 90.25 $\pm$ 2.5 & 94.82 $\pm$ 1.4 & 9.68 $\pm$ 2.5  \\ 
$a_4$  & 100.0 $\pm$ 0.0 & 93.76 $\pm$ 0.8 & 96.78 $\pm$ 0.4 & 6.24 $\pm$ 0.8  \\ 
$\sum$ & 99.99 $\pm$ 0.0 & 95.09 $\pm$ 0.7 & 97.48 $\pm$ 0.4 & 4.89 $\pm$ 0.7  \\ 
\noalign{\smallskip}\hline\noalign{\smallskip}
\end{tabular}
\end{table}

\begin{table}
\caption{GWR Performance with Objects with $a_T=0.95$}
\label{tab:gwr_predicta95}
\begin{tabular}{lllll}
\noalign{\smallskip}\hline\noalign{\smallskip}
       & Precision (\%) & Recall (\%) & F1 (\%)  & Misses (\%) \\
\hline\noalign{\smallskip}
       & 30 epochs & IoU:0.754 $\pm$ 0.002 & & \\
\noalign{\smallskip}\hline\noalign{\smallskip}
$a_1$  & 100.0 $\pm$ 0.0 & 100.0 $\pm$ 0.0 & 100.0 $\pm$ 0.0  & 0.0 $\pm$ 0.0  \\ 
$a_2$  & 100.0 $\pm$ 0.0 & 99.19 $\pm$ 0.3 & 99.6 $\pm$ 0.1   & 0.81 $\pm$ 0.3 \\ 
$a_3$  & 99.79 $\pm$ 0.2 & 97.33 $\pm$ 0.4 & 98.54 $\pm$ 0.1  & 2.47 $\pm$ 0.5 \\ 
$a_4$  & 100.0 $\pm$ 0.0 & 98.0 $\pm$ 0.4  & 98.99 $\pm$ 0.2  & 2.0 $\pm$ 0.4  \\ 
$\sum$ & 99.96 $\pm$ 0.0 & 98.54 $\pm$ 0.2 & 99.2 $\pm$ 0.1   & 1.51 $\pm$ 0.3 \\ 
\noalign{\smallskip}\hline\noalign{\smallskip}
       & 50 epochs & IoU:0.756 $\pm$ 0.001 & & \\
\noalign{\smallskip}\hline\noalign{\smallskip}
$a_1$  & 100.0 $\pm$ 0.0 & 100.0 $\pm$ 0.0 & 100.0 $\pm$ 0.0  & 0.0 $\pm$ 0.0  \\ 
$a_2$  & 100.0 $\pm$ 0.0 & 99.14 $\pm$ 0.6 & 99.57 $\pm$ 0.3  & 0.86 $\pm$ 0.6 \\ 
$a_3$  & 100.0 $\pm$ 0.0 & 96.46 $\pm$ 0.7 & 98.2 $\pm$ 0.4   & 3.54 $\pm$ 0.7 \\ 
$a_4$  & 100.0 $\pm$ 0.0 & 98.16 $\pm$ 0.7 & 99.07 $\pm$ 0.4  & 1.84 $\pm$ 0.7 \\ 
$\sum$ & 100.0 $\pm$ 0.0 & 98.33 $\pm$ 0.3 & 99. 16 $\pm$ 0.1 & 1.67 $\pm$ 0.3 \\ 
\noalign{\smallskip}\hline\noalign{\smallskip}
       & 100 epochs & IoU:0.762 $\pm$ 0.001 & & \\
\noalign{\smallskip}\hline\noalign{\smallskip}
$a_1$  & 100.0 $\pm$ 0.0 & 100.0 $\pm$ 0.0 & 100.0 $\pm$ 0.0  & 0.0 $\pm$ 0.0  \\ 
$a_2$  & 100.0 $\pm$ 0.0 & 99.34 $\pm$ 0.3 & 99.67 $\pm$ 0.2  & 0.66 $\pm$ 0.3 \\ 
$a_3$  & 100.0 $\pm$ 0.0 & 97.08 $\pm$ 0.2 & 98.52 $\pm$ 0.1  & 2.92 $\pm$ 0.2 \\ 
$a_4$  & 100.0 $\pm$ 0.0 & 98.08 $\pm$ 0.1 & 99.03 $\pm$ 0.1  & 1.92 $\pm$ 0.1 \\ 
$\sum$ & 100.0 $\pm$ 0.0 & 98.47 $\pm$ 0.1 & 99.23 $\pm$ 0.1  & 1.53 $\pm$ 0.1 \\ 
\noalign{\smallskip}\hline\noalign{\smallskip}
\end{tabular}
\end{table}

The evaluation of the prediction with GWRs coupled with ambiguity and noise detection shows considerable improvements for class $a_1$, where both precision and recall achieve 100\% across all networks and parameter values $a_T$. Due to the ambiguity detection, expressed as \textit{correctly detected ambiguity} (CDA) in tables \ref{tab:gwr_cda85}, \ref{tab:gwr_cda90}, and \ref{tab:gwr_cda95}, we can show that this extra step caused a decrease in \textit{misses}, a crucially negative factor limiting the recall abilities as demonstrated above. Also, the number of \textit{misses} could be improved from 12.77\% to 5.4\% for 30 epochs and $a_T=0.85$ and even to 1.67\% for $a_T=0.90$, positively influencing the recall and F1 score. 

Regarding the CDA, it becomes evident, overall, that the performance is reasonably good except for class $a_2$. We explain this by the fact that objects are closer together and thus can be captured more easily. Noticeably, no noise computed as (falsely detected noise, FDN) is detected across all networks.

However, our results also reveal a decrease in performance for the ambiguity classes $a_3$ and $a_4$. While the differences of the IoU values are rather small (e.g. a drop from 0.725 to 0.713 for 30 epochs and $a_T=0.90$), the \textit{misses} increased considerably. For instance, for 50 epochs and $a_T=0.90$ in $a_3$ this number more than doubled from 4.08\% to 10.37\%, and similarly for $a_4$ from 2.47\% to 6.81\%. This effect is reduced when considering $a_t=0.95$. Our findings show the sensitivity of an integrated object detection for pointing gestures, rendering their recognition non-trivial. When testing our system to new object arrangements, the divergence between the manually annotated bounding boxes and the predictions returned by the GWR is even more noticeable with increasing levels of ambiguities. However, we want to emphasize that our object detection implementation is very simple, yet the GWR predictions yield good results for our gesture scenario.  


\begin{table}
\caption{CDA and FDN, $a_T=0.85$}
\label{tab:gwr_cda85}
\begin{tabular}{lll}
 & CDA (\%) & FDN (\%) \\
\hline\noalign{\smallskip}
 & 30 epochs  & \\
\noalign{\smallskip}\hline\noalign{\smallskip}
$a_1$ & 100 $\pm$ 0.0 & 0.0 $\pm$ 0.0 \\ 
$a_2$ & 70.88 $\pm$ 1.8 & 0.0 $\pm$ 0.0 \\ 
$a_3$ & 97.12 $\pm$ 1.4 & 0.0 $\pm$ 0.0 \\ 
$a_4$ & 93.89 $\pm$ 0.8 & 0.0 $\pm$ 0.0 \\ 
$\sum$ & 90.93 $\pm$ 0.2 & 0.0 $\pm$ 0.0 \\ 
\hline\noalign{\smallskip}
 & 50 epochs  & \\
\noalign{\smallskip}\hline\noalign{\smallskip}
$a_1$ & 99.59 $\pm$ 0.6 & 0.0 $\pm$ 0.0 \\ 
$a_2$ & 70.51 $\pm$ 3.3 & 0.0 $\pm$ 0.0 \\ 
$a_3$ & 97.69 $\pm$ 0.7 & 0.0 $\pm$ 0.0 \\ 
$a_4$ & 93.53 $\pm$ 1.6 & 0.0 $\pm$ 0.0 \\ 
$\sum$ & 90.73 $\pm$ 1.0 & 0.0 $\pm$ 0.0 \\
\hline\noalign{\smallskip}
 & 100 epochs  & \\
\noalign{\smallskip}\hline\noalign{\smallskip}
$a_1$ & 99.38 $\pm$ 0.9 & 0.0 $\pm$ 0.0 \\ 
$a_2$ & 69.39 $\pm$ 0.4 & 0.0 $\pm$ 0.0 \\ 
$a_3$ & 98.85 $\pm$ 1.3 & 0.0 $\pm$ 0.0 \\ 
$a_4$ & 95.05 $\pm$ 0.2 & 0.0 $\pm$ 0.0 \\ 
$\sum$ & 91.63 $\pm$ 0.2 & 0.0 $\pm$ 0.0 \\
\noalign{\smallskip}\hline\noalign{\smallskip}
\end{tabular}
\end{table}


\begin{table}
\caption{CDA and FDN, $a_T=0.90$}
\label{tab:gwr_cda90}
\begin{tabular}{lll}
 & CDA (\%) & FDN (\%) \\
\hline\noalign{\smallskip}
 & 30 epochs  & \\
\noalign{\smallskip}\hline
$a_1$ & 100 $\pm$ 0.0 & 0.0 $\pm$ 0.0 \\ 
$a_2$ & 72.8 $\pm$ 1.5 & 0.0 $\pm$ 0.0 \\
$a_3$ & 99.41 $\pm$ 0.5 & 0.0 $\pm$ 0.0 \\ 
$a_4$ & 94.63 $\pm$ 1.1 & 0.0 $\pm$ 0.0 \\ 
$\sum$ & 92.06 $\pm$ 0.8 & 0.0 $\pm$ 0.0 \\
\noalign{\smallskip}\hline
 & 50 epochs  & \\
\noalign{\smallskip}\hline
$a_1$ & 100 $\pm$ 0.0 & 0.0 $\pm$ 0.0 \\ 
$a_2$ & 72.52 $\pm$ 0.3 & 0.0 $\pm$ 0.0 \\
$a_3$ & 98.65 $\pm$ 1.0 & 0.0 $\pm$ 0.0 \\ 
$a_4$ & 96.48 $\pm$ 0.9 & 0.0 $\pm$ 0.0 \\ 
$\sum$ & 92.71 $\pm$ 0.4 & 0.0 $\pm$ 0.0 \\
\noalign{\smallskip}\hline
 & 100 epochs  & \\
\noalign{\smallskip}\hline
$a_1$ & 100 $\pm$ 0.0 & 0.0 $\pm$ 0.0 \\ 
$a_2$ & 72.36 $\pm$ 1.6 & 0.0 $\pm$ 0.0 \\ 
$a_3$ & 98.68 $\pm$ 0.4 & 0.0 $\pm$ 0.0 \\ 
$a_4$ & 95.47 $\pm$ 1.5 & 0.0 $\pm$ 0.0 \\ 
$\sum$ & 92.22 $\pm$ 0.8 & 0.0 $\pm$ 0.0 \\
\noalign{\smallskip}\hline
\end{tabular}
\end{table}


\begin{table}
\caption{CDA and FDN, $a_T=0.95$}
\label{tab:gwr_cda95}\begin{tabular}{lll}
\hline\noalign{\smallskip}
 & 30 epochs  & \\
\noalign{\smallskip}\hline
$a_1$ & 100 $\pm$ 0.0 & 0.0 $\pm$ 0.0 \\ 
$a_2$ & 85.73 $\pm$ 1.4 & 0.0 $\pm$ 0.0 \\ 
$a_3$ & 99.8 $\pm$ 0.3 & 0.0 $\pm$ 0.0 \\ 
$a_4$ & 94.07 $\pm$ 1.2 & 0.0 $\pm$ 0.0 \\ 
$\sum$ & 94.45 $\pm$ 0.3 & 0.0 $\pm$ 0.0 \\ 
\hline\noalign{\smallskip}
 & 50 epochs  & \\
\hline\noalign{\smallskip}
$a_1$ & 100 $\pm$ 0.0 & 0.0 $\pm$ 0.0 \\ 
$a_2$ & 80.82 $\pm$ 0.9 & 0.0 $\pm$ 0.0 \\
$a_3$ & 100 $\pm$ 0.0 & 0.0 $\pm$ 0.0 \\ 
$a_4$ & 94.49 $\pm$ 1.3 & 0.0 $\pm$ 0.0 \\ 
$\sum$ & 93.7 $\pm$ 0.7 & 0.0 $\pm$ 0.0 \\ 
\noalign{\smallskip}\hline\noalign{\smallskip}
 & 100 epochs  & \\
\hline\noalign{\smallskip}
$a_1$ & 100 $\pm$ 0.0 & 0.0 $\pm$ 0.0 \\ 
$a_2$ & 78.4 $\pm$ 1.0 & 0.0 $\pm$ 0.0 \\ 
$a_3$ & 100 $\pm$ 0.0 & 0.0 $\pm$ 0.0 \\ 
$a_4$ & 96.83 $\pm$ 0.5 & 0.0 $\pm$ 0.0 \\ 
$\sum$ & 94.27 $\pm$ 0.0 & 0.0 $\pm$ 0.0 \\ 
\noalign{\smallskip}\hline
\end{tabular}
\end{table}

\subsection{Comparative Evaluation}

Our evaluation results for both the computer vision approach and our novel model proposal using GWRs show good performance for ambiguity classes $a_1$ and $a_2$, achieving even 100\% precision. With an increasing level of ambiguity, the computer vision approach fails to capture the pointing-object associations, expressed in a high value of \textit{misses} of 22.96\% compared to 0.62\% to 9.31\% when evaluated individually, and 1.84\% to 14.56\% for the GWR prediction. As a result, the GWRs also show better performance in terms of the recall. From our experiments, especially regarding ambiguity class $a_4$, we can conclude that the computer vision approach is highly sensitive to the environmental lab conditions, suffering from image artifacts due to lighting conditions or other confounding factors. In addition, our implemented ambiguity detection shows support for the prediction process, yielding 100\% precision among all networks. Finally, our proposed approach modelling pointing gestures with GWR provides a more controlled way for recognizing a pointing gesture: the computer vision approach always extrapolates a certain pointing position and would always target to match a corresponding object in the scene. Thus, in an extreme case, pointing to the ceiling or to another wrong direction would always result in a pointing intention and, therefore, the next object would be aimed for to establish the pointing-object association. In contrast, the GWR allows us to model a set of pointing directions beforehand, independent of any markers on the table. The flexibility of designing a certain topology or range for target directions yields a more accurate prediction of the pointing and excludes any hand movements, may they be random or simply not in the required scenario range.

\section{Discussion and Future Work}

In the present paper we proposed a novel modeling approach for the implementation of a flexible yet robust interface for pointing gesture in HRI scenarios. We designed experiments with distinct object arrangements for which we defined different ambiguity classes. We explained our scenario and experimental requirements to facilitate the reproduction of our procedures. The NICO humanoid robot was selected due to current developments on this platform. However, we used only the on-board RGB camera to be compatible with a large range of robotic platforms that have RGB cameras or simply standalone camera devices. To use the GWR to model the gesture distribution, we introduced a specific labeling function and designed a method to evaluate different labels returned by a trained network.
For a fair evaluation comparison, we chose standard computer vision approaches as a performance baseline. We focused on the gestures and the interplay with the different ambiguities, thus we restricted the objects to be simple, differently colored cubes. Our comparative evaluation showed reasonable performance of our baseline and superior performance of the GWR. Even for a rather small number of epochs (namely 30 and an $a_T=90$) in the network generation, we achieved 93.94\% correctly classified gestures. Our results confirmed the hypothesis that GWRs are effective to model the topology of the gestures employed in our scenario, where the strength lies in their flexibility of learning any range of pointing directions. Therefore, our approach does not need to handle outliers like random movements or unintentional gestures.

Our HRI pointing scenario fits well into the current demand for assisting robotic applications and social robots \cite{Breaz_04}\cite{Canal_16}. However, our approach differs from approaches using a complete motion capture system \cite{Hagit_07} or employing primarily depth sensor information \cite{Cosgu_15}\cite{Shukl_15}. Although typical issues like background removal and separation of distinct objects in the scene may be facilitated, corresponding devices inherently suffer from sensor noise and unreliable tracking of the skeleton returned by standard libraries. In addition, pointing gestures are usually captured by a robot sitting in close proximity to the target scenario, e.g., on a chair or sofa close to a table. Thus, the camera tracks only the upper body part, which makes an initial full-body calibration unnecessary and is even a very unnatural way of interaction. 

Using the GWR seems counter-intuitive as we need an additional labeling procedure known from supervised learning. In our context, this method allows us to capture the distribution of the gestures and, as such, to model the topology or range of those gestures. As a result, we can ``discretize" the inherently continuous gesture space without any markers for predefined areas, prespecified pointing directions or other artificially created restrictions. Our performance evaluations demonstrate the robustness of our approach both in terms of correct pointing gestures when ambiguity is present as well as filtering out noise. Although Shukla et al. \cite{Shukl_15} showed a good performance of 99.4\%, the accuracy drops significantly when regulating their introduced factor $t$ to be more conservative. Also, our model showed more stable behaviour even for ambiguities for objects in close proximity in contrast to Cosgun et al. \cite{Cosgu_15}, where distances were only allowed to be 2cm away and only the presence of ambiguity was returned.
Many improvements may increase the capability of our baseline approach, but we demonstrated the benefit of the GWR to model pointing gestures, which we see as our strongest contribution to this research area.

Our present implementation is integrated into current research of the NICO robot \cite{Kerze_17} and a first step coupling the system for modeling more complex behaviour. Thus, the current development stage is limited in some aspects, for which we would like to point out suggestions for future improvements and extensions.
Regarding the feature selection, we could not find evidence of superior performance using optical flow (left out for brevity reason) in contrast to the study presented by Y. Nagai \cite{Nagai_05a}, which computationally underpinned the significance of movements in developmental learning \cite{Moore_97}\cite{Astor_19}. However, our limiting factor in the study might be the lack of any integration of face features and head movements like gaze. Therefore, we suggest to extend our system with facial expressions and gaze features, where flow vectors assigning the direction of interest may be of substantial benefit. 

Similarly, we assumed to a one-person scenario \cite{Nagai_05} due to the desired evaluation of object ambiguities. However, evaluating our approach using a dataset providing hand poses from multiple views \cite{Shukl_16} and integrating it in our current system would show further evidence of the robustness of our work. We hope that providing the code along with the paper could stimulate the reproduction and the establishment of a baseline deictic gesture dataset, i.e. including a set of pointing gestures but also other gestures in the general set of deictic gestures \cite{Saupp_14}. 

To make a relevant contribution to social robotics, it is essential to integrate robots as active communication partners in HRI scenarios, e.g. by adding a dialogue system \cite{Hagit_07} or implementing a feedback loop asking the human for help. Also, another important extension to the present architecture is the integration of real-world objects, possibly even connected to the recognition of object affordances. This would supply necessary object information that may help the robot both in the concrete object recognition and in a subsequent grasping task to avoid hardware load, e.g. when lifting an object is impossible due to constraints in the robot design or in the spatial arrangement. The latter aspect would play an important role when also extending the arrangements to object stacks or when adding partial occlusions in the scene.

Finally, it is necessary to evaluate HRI scenarios regarding the subjects' experience with the robot \cite{Saupp_14}\cite{Canal_16}. The rather poor performance of the pointing gestures as shown in the study conducted by Saupp\'{e} and Mutlu \cite{Saupp_14} is, in this context, another interesting result worth to investigate further.

\section{Conclusion}
Our current work contributes to the research area of deictic gestures, specifically pointing gestures, in HRI scenarios by introducing the GWR as a method to model an inherently continuous space, which renders the usage of specific markers or predefined positions unnecessary. Our approach relies on standard camera images and can thus be easily used and integrated into other systems or robotic platforms.
The validity of our approach is underpinned by a thorough evaluation of a set of pointing gestures, performed under easy and difficult conditions introduced by our notion of ambiguity classes. The overall superior performance of our approach is a promising framework for future extensions of our system for HRI research and applications.

\section*{Acknowledgements}
We would like to thank German I. Parisi for the support in the GWR setup.
The authors gratefully acknowledge partial support from the German Research Foundation DFG under project CML (TRR 169).


%
%

\end{document}